\documentclass{article}

\usepackage[final]{neurips_2025}

\usepackage[utf8]{inputenc} 
\usepackage[T1]{fontenc}    
\usepackage{hyperref}       
\usepackage{url}            
\usepackage{booktabs}       
\usepackage{amsfonts}       
\usepackage{nicefrac}       
\usepackage{microtype}      
\usepackage{xcolor}         
\usepackage{graphicx}
\usepackage{amsmath}
\usepackage{amsfonts}
\usepackage{multirow}
\usepackage{subfigure}
\usepackage{pgfplots}
\usepackage{xspace}
\usepackage{enumitem}
\usepackage{wrapfig}
\usepackage{bbding}
\usepackage{colortbl}
\usepackage{adjustbox}
\usepackage{caption}

\title{Teaching Language Models to Evolve with Users: Dynamic Profile Modeling for Personalized Alignment}

\newcommand*\samethanks[1][\value{footnote}]{\footnotemark[#1]}

\author{%
  Weixiang Zhao\textsuperscript{1}\thanks{\ \ Equal contribution}\, , Xingyu Sui\textsuperscript{1}\samethanks \, , Yulin Hu\textsuperscript{1}\samethanks\, , Jiahe Guo\textsuperscript{1} \\ \textbf{Haixiao Liu\textsuperscript{2}}, \textbf{Biye Li\textsuperscript{2}}, \textbf{Yanyan Zhao\textsuperscript{1}}\thanks{\ \ Corresponding author}\, , \textbf{Bing Qin\textsuperscript{1}}, \textbf{Ting Liu\textsuperscript{1}}\\
  \textsuperscript{1}Harbin Institute of Technology, \textsuperscript{2}Du Xiaoman Financial\\
  \texttt{\{wxzhao,jhguo,yyzhao\}@ir.hit.edu.cn} \\
}

\begin{document}

\maketitle

\begin{abstract}
  Personalized alignment is essential for enabling large language models (LLMs) to engage effectively in user-centric dialogue. While recent prompt-based and offline optimization methods offer preliminary solutions, they fall short in cold-start scenarios and long-term personalization due to their inherently static and shallow designs. In this work, we introduce the Reinforcement Learning for Personalized Alignment (RLPA) framework, in which an LLM interacts with a simulated user model to iteratively infer and refine user profiles through dialogue. The training process is guided by a dual-level reward structure: the Profile Reward encourages accurate construction of user representations, while the Response Reward incentivizes generation of responses consistent with the inferred profile. We instantiate RLPA by fine-tuning Qwen-2.5-3B-Instruct, resulting in Qwen-RLPA, which achieves state-of-the-art performance in personalized dialogue. Empirical evaluations demonstrate that Qwen-RLPA consistently outperforms prompting and offline fine-tuning baselines, and even surpasses advanced commercial models such as Claude-3.5 and GPT-4o. Further analysis highlights Qwen-RLPA's robustness in reconciling conflicting user preferences, sustaining long-term personalization and delivering more efficient inference compared to recent reasoning-focused LLMs. These results emphasize the potential of dynamic profile inference as a more effective paradigm for building personalized dialogue systems. Our code is available at: \href{https://github.com/XingYuSSS
/RLPA}{https://github.com/XingYuSSS/RLPA}.
\end{abstract}

\section{Introduction}
In recent years, aligning large language models (LLMs) with human values and intentions has become a crucial prerequisite for deploying them in interactive applications \citep{askell2021general,bai2022constitutional,zhao2023survey}. Alignment techniques---such as instruction tuning and reinforcement learning from human feedback (RLHF)---have significantly improved the helpfulness and harmlessness of model outputs by aligning them with broadly shared human preferences \citep{ouyang2022training,ji2023ai,shen2023large}. This form of general alignment enables LLMs to follow instructions, avoid unsafe content, and exhibit socially acceptable behavior across a wide range of users.

However, such a one-size-fits-all alignment paradigm fails to account for the diversity of individual user needs, goals, and interaction styles \citep{sorensen2024position,kirk2024benefits,jiang2024peek}. As a result, it limits the model's capacity to engage in personalized communication, which is essential for domains such as long-term dialogue, personalized education, and user-centered decision support \citep{zhang2024personalization,tseng2024two,wu2024personalized,salemi2024lamp,liu2025survey}.

While personalized alignment holds great promise for enhancing user experience, achieving it in practice remains a significant challenge. A core difficulty stems from the inherently dynamic and evolving nature of personalization \citep{shi2024wildfeedback,jiang2025know}. In particular, effective personalized alignment requires the model to continuously infer and adapt to user-specific attributes---such as preferences, goals, and beliefs---throughout the interaction. In real-world settings, the model often encounters a \emph{cold-start} scenario, where little or no prior user information is available. As a result, it must dynamically construct and refine user representations based solely on the dialogue context, enabling long-term adaptation and the development of coherent user profiles across extended conversations \citep{zhao2025llms,xie2025survey}.

\begin{figure}
  \centering
  \includegraphics[width=\linewidth]{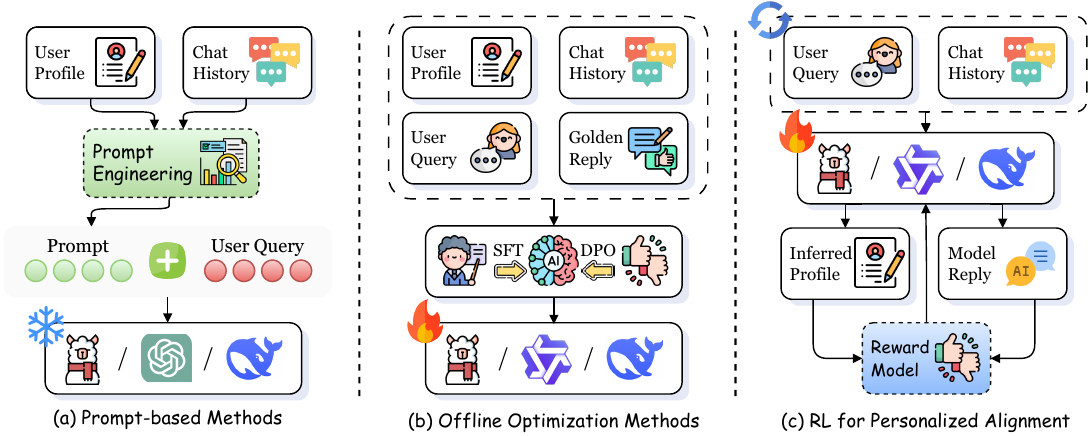}
  \caption{Comparison of personalized alignment paradigms. (a) Prompt-based methods inject user profiles and chat history into the prompt at inference time. (b) Offline optimization methods (e.g., SFT, DPO) rely on static data with predefined replies. and (c) Our proposed RLPA framework models personalization as a multi-turn interactive process optimized via reinforcement learning.}
  \label{fig:motivation}
\end{figure}

While recent efforts have sought to address these challenges of personalization, they remain inadequate in meeting the demands of dynamic and adaptive user needs. As shown in Figure \ref{fig:motivation}(a), prompt-based methods---such as profile-augmented prompting \citep{wang2023chain,richardson2023integrating,pandey2024cos,li2024learning} and retrieval-augmented prompting \citep{zhang2024llm,li2024hello,salemi2024optimization,zhuang2024hydra}---typically provide only superficial personalization, relying on static templates to inject user-specific information. This approach limits both the flexibility and expressiveness of model outputs and hampers the integration of long-term user knowledge due to the constraints of the context window \citep{liu2025survey}. Meanwhile, offline optimization methods in Figure \ref{fig:motivation}(b), such as supervised fine-tuning (SFT) \citep{ouyang2022training,clarke2024peft,tan2024personalized,peng2024pocketllm} and direct preference optimization (DPO) \citep{rafailov2023direct,jang2023personalized,kirk2024prism,zollo2024personalllm,chen2025deeper}, require large-scale labeled datasets, making them impractical in cold-start scenarios where user data is scarce or unavailable. Moreover, these methods tend to generalize poorly across users due to their static nature \citep{xu2024dpo,lin2024limited,chu2025sft}, rendering them inflexible for real-time adaptation during interactions.

To address these limitations, we formulate the task of personalized alignment as a multi-turn Markov Decision Process (MDP), where the model interacts with a user over multiple dialogue turns to infer and adapt to personalized preferences. To solve this MDP, we introduce the \underline{\textbf{R}}einforcement \underline{\textbf{L}}earning for \underline{\textbf{P}}ersonalized \underline{\textbf{A}}lignment (\textbf{RLPA}) framework, in which the model learns through online interaction with a simulated user model that provides dynamic and consistent user behavior. Specifically, we design a two-level reward mechanism to supervise the learning process: a Profile Reward guides the model to extract and update user-specific attributes, from the dialogue history, enabling the construction of dynamic user profiles. In parallel, a Response Reward encourages the model to generate responses aligned with the inferred profile, enhancing personalization quality. Both rewards are provided at every dialogue turn, enabling immediate feedback and continuous adaptation. Through this reward-driven multi-turn interaction, the model progressively learns to infer, maintain, and leverage user profiles in a manner well-suited to cold-start and dynamically evolving user scenarios.

We fine-tune the Qwen-2.5-3B-Instruct model \citep{yang2024qwen2} using our proposed RLPA framework, resulting in the Qwen-RLPA model. Experimental results show that Qwen-RLPA substantially outperforms both prompt-based and offline optimization baselines in terms of personalization quality. Notably, it surpasses leading proprietary systems such as Claude-3.5 and DeepSeek-V3, and achieves performance on par with GPT-4o. Further analysis indicates that Qwen-RLPA is capable of sustaining coherent, personalized responses throughout extended long-term dialogues, effectively resolving preference conflicts and dynamically adapting its behavior. Moreover, when compared against recent state-of-the-art reasoning LLMs, including DeepSeek-R1 \citep{guo2025deepseek} and OpenAI-o3 \citep{openai2025o3blog}, Qwen-RLPA delivers superior performance with significantly greater inference efficiency. These results highlight the potential of dynamically constructed user profiles as a more appropriate and effective reasoning paradigm for personalized dialogue systems.

In summary, our work makes the following contributions:
\begin{itemize}[leftmargin=*]
    \item We conceptualize the task of personalized dialogue alignment as a multi-turn Markov Decision Process, capturing the dynamic and evolving nature of user preference modeling under cold-start and real-time adaptation scenarios.
    \item We propose the RLPA framework, which trains LLMs via interaction with a simulated user using a dual-level reward mechanism at every dialogue turn.
    \item We fine-tune Qwen-2.5-3B-Instruct using RLPA and show that the resulting Qwen-RLPA model significantly outperforms both open and closed-source baselines in personalization quality, long-term coherence, demonstrating its effectiveness of dynamic personalization.
\end{itemize}

\section{Preliminaries}

In this section, we first formalize the personalized alignment in dialogue as a multi-turn Markov Decision Process (\S\ref{subsec:mdp}). We then introduce the reinforcement learning as the optimization paradigm for learning effective personalization policies (\S\ref{subsec:rl}).

\subsection{Personalized Alignment as a Multi-Turn Markov Decision Process}
\label{subsec:mdp}

We model personalized alignment as a multi-turn MDP, defined by the tuple $(\mathcal{S}, \mathcal{A}, \mathcal{T}, \mathcal{R}, \gamma)$. The agent (a language model) interacts with a user to incrementally infer and adapt to user-specific attributes throughout the dialogue.
\begin{itemize}[leftmargin=*]
	\item \textbf{State} $\mathcal{S}$: At turn $t$, the state $s_t = \{u_1, r_1, \dots, u_t\}$ contains the dialogue history so far, enabling progressive inference of latent user profiles.
	\item \textbf{Action} $\mathcal{A}$: The action $a_t$ corresponds to the generated response $r_t$ at turn $t$.
	\item \textbf{Transition} $\mathcal{T}$: Given $(s_t, r_t)$, the environment (simulated user) returns the next user utterance $u_{t+1}$, updating the state to $s_{t+1}$.
	\item \textbf{Reward} $\mathcal{R}$: At each turn, the agent receives a reward composed of two components: a profile reward that evaluates profile inference accuracy, and a response reward that measures alignment between the response and inferred profile.
	\item \textbf{Discount Factor} $\gamma \in [0,1]$: Balances immediate and long-term rewards.
\end{itemize}

This formulation captures the sequential and adaptive nature of personalized dialogue, where user profiles must be inferred and refined dynamically across turns.

\subsection{Reinforcement Learning for Policy Optimization}
\label{subsec:rl}

Given the MDP setup, we aim to learn a dialogue policy $\pi(a_t \mid s_t)$ that maximizes the expected cumulative reward:
\begin{equation}
J(\pi) = \mathbb{E}_\pi \left[ \sum_{t=1}^{T} \gamma^{t-1} R_t \right]
\end{equation}
We adopt reinforcement learning (RL) as our personalized alignment training paradigm, enabling the model to learn from interaction with simulated users rather than static labeled data. At each turn, the model observes a current state $s_t$, generates a response $a_t$, receives a reward $R_t$, and updates its policy based on long-term outcomes.

Unlike supervised learning, RL supports delayed and cumulative feedback, allowing the model to optimize two interdependent goals: (1) accurately tracking user preferences, and (2) generating coherent and personalized responses accordingly. This makes RL particularly suited for building adaptive and user-centric dialogue agents in the cold-start scenario.

\begin{figure}
  \centering
  \includegraphics[width=\linewidth]{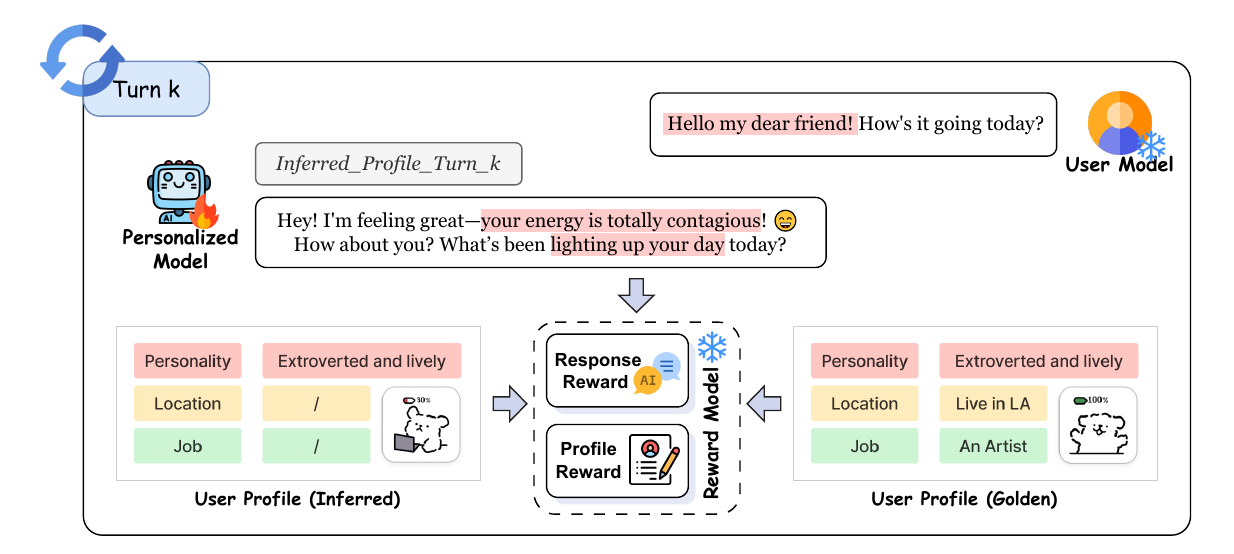}
  \caption{Overview of the RLPA framework. The policy model interacts with a simulated user whose responses are conditioned on a pre-defined profile. At each turn, the model generates a response and an inferred profile, which are evaluated by two reward functions: the Profile Reward supervises the accuracy of user modeling, while the Response Reward assesses personalization quality. The combined reward is used to optimize the model policy via reinforcement learning.}
  \label{fig:method}
\end{figure}

\section{Method: Reinforcement Learning for Personalized Alignment (RLPA)}

This section introduces our Reinforcement Learning for Personalized Alignment (RLPA) framework, which trains LLMs to dynamically infer and respond to user-specific preferences through interaction, guided by reward signals that evaluate both profile inference and personalized response quality.

Specifically, as shown in Figure \ref{fig:method}, RLPA consists of the following three core components:
\begin{itemize}[leftmargin=*]
	\item Simulated User Design (\S\ref{subsec:user}). We construct a controllable user simulator that provides realistic, profile-grounded responses and exposes latent user attributes across dialogue turns.
	\item Profile Reward Function Design (\S\ref{subsec:profile}). We define a reward signal that quantifies how accurately the model captures and updates user-specific information based on observed interactions.
	\item Response Reward Function Design (\S\ref{subsec:response}). We develop a mechanism to evaluate the model’s responses for personalization fidelity, ensuring they align with the current inferred profile.
\end{itemize}

\subsection{Simulated User Design}
\label{subsec:user}

To facilitate scalable and controllable training, we construct the simulated user model that interacts with the dialogue agent and provides consistent, profile-grounded responses. This setup allows the model to learn personalization strategies through reinforcement learning.

Each simulated user is initialized with a given profile $\mathcal{P} = \{p_1, p_2, \dots, p_n\}$, where each $p_i$ denotes a user attribute such as preferences, personality traits, or goals. These attributes are embedded into the system prompt of the user model, which conditions its behavior throughout the dialogue \citep{brown2020language,kojima2022large,DBLP:conf/nips/ZhengWZN0C24,zhao2025beware}.

At each turn $t$, given the dialogue history $H_t = \{u_1, r_1, \dots, u_{t-1}, r_{t-1}\}$ and the model response $r_t$, the simulator generates the next user utterance $u_{t+1}$ via its behavior policy $\pi_u(u_{t+1} \mid H_t, \mathcal{P})$. This setup ensures the simulated user behaves coherently and reflects the underlying profile across turns. The simulated user satisfies two key properties: (1) Profile Groundedness: Responses are consistently shaped by the injected profile $\mathcal{P}$, providing inference cues for the dialogue agent. Importantly, the user model is instructed to reveal profile information gradually over turns, rather than disclosing all attributes at once. This incremental exposure encourages the dialogue agent to accumulate and refine its understanding of the user profile through multi-turn reasoning. (2) Behavioral Consistency: The user exhibits stable preferences and conversational style over time, allowing the model to benefit from cumulative reasoning across multiple turns.

To assess whether the user model satisfies the desired properties, we conduct human evaluations using various base models as user simulators. The results reveal notable differences in profile fidelity and behavioral consistency across models. Balancing performance with computational cost, we ultimately select GPT-4o-mini as the user model for all subsequent experiments. Comprehensive evaluation results and comparisons are provided in Appendix \ref{app:user_eval}.

\subsection{Profile Reward Function Design}
\label{subsec:profile}

The Profile Reward is designed to guide the model in accurately inferring and maintaining a user profile $\hat{\mathcal{P}}$ throughout multi-turn dialogues. As user attributes are not explicitly observable, the model must implicitly infer them from user utterances and continuously update its internal representation.

To enable structured inference and efficient matching, we represent the user profile $\mathcal{P}$ using a slot-value format, where each slot $p_i$ denotes a predefined attribute category paired with a specific value (see Figure \ref{fig:example_slot} in Appendix \ref{app:eval_data} for detailed examples). At each dialogue turn $t$, the model is trained to produce its current estimate $\hat{\mathcal{P}}_t = {\hat{p}_{1,t}, \dots, \hat{p}_{n,t}}$ based on the dialogue history up to that point. This structured representation supports explicit supervision by allowing direct comparison between the predicted and ground-truth profile slots. The reward is calculated using a slot-wise matching score:
\begin{equation}
\text{Precision}_t = \frac{|\hat{\mathcal{P}}_t \cap \mathcal{P}|}{|\hat{\mathcal{P}}_t|}, \quad
\text{Recall}_t = \frac{|\hat{\mathcal{P}}_t \cap \mathcal{P}|}{|\mathcal{P}|}
\end{equation}
\begin{equation}
R^{\text{profile}}_t = \frac{2 \cdot \text{Precision}_t \cdot \text{Recall}_t}{\text{Precision}_t + \text{Recall}_t} = \frac{2 \cdot |\hat{\mathcal{P}}_t \cap \mathcal{P}|}{|\hat{\mathcal{P}}_t| + |\mathcal{P}|}
\label{eq:profile_reward}
\end{equation}
This reward formulation encourages the model to incrementally infer accurate and complete user profiles by rewarding overlapping slot-value matches while penalizing omissions and incorrect predictions. As such, \( R^{\text{profile}}_t \) provides a structured and interpretable training signal aligned with the goal of profile tracking throughout the dialogue.



\subsection{Response Reward Function Design}
\label{subsec:response}

While the profile reward supervises user modeling, the response reward ensures that generated responses faithfully reflect the inferred profile. At each dialogue turn t, an external reward model evaluates the alignment between the response $r_t$ and the inferred profile $\hat{\mathcal{P}}_t$, focusing on personalization rather than surface lexical similarity.

The reward model outputs a scalar score $R^{\text{response}}_t \in [0,1]$, based on four core dimensions: preference expression, style consistency, goal alignment, and persona coherence. To enforce output quality, we further require the response to satisfy five binary criteria—Naturalness (N), Relevance (R), Logical consistency (L), Engagement (G), and Informativeness (F)—and compute the final reward as:
\begin{equation}\small
R^{\text{response}}_t = N \cdot R \cdot L \cdot G \cdot F
\label{eq:response_reward}
\end{equation}
This strict formulation rewards only fully satisfactory responses across all aspects.

To select a reliable reward model, we benchmark several LLMs (GPT-4o, DeepSeek-V3) by comparing their scoring consistency with human judgments. GPT-4o yields the highest agreement, but for efficiency, we adopt GPT-4o-mini during RL training. Full results are shown in Appendix \ref{app:response_reward_model}.

\subsection{Training with Proximal Policy Optimization (PPO)}

To perform optimization in RLPA, we adopt Proximal Policy Optimization (PPO) \citep{schulman2017proximal}, a widely used policy gradient algorithm. At each dialogue turn $t$, the model receives a combined reward signal that supervises both user modeling and personalized response generation:
\begin{equation}
R_t = R^{\text{profile}}_t + R^{\text{response}}_t
\end{equation}
This turn-level reward encourages the model to continuously infer, refine, and utilize user profiles throughout the interaction. The full PPO optimization objective are provided in Appendix~\ref{app:ppo}.

\section{Experiments}

\subsection{Experimental Setup}

\paragraph{Models \& Training Data} We instantiate our RLPA framework using the Qwen-2.5-3B-Instruct \citep{yang2024qwen2}. For the user simulator, we adopt GPT-4o-mini, selected based on our human evaluation study (see Appendix \ref{app:user_eval}). The reward model is also implemented using GPT-4o-mini and prompted to assess response alignment with user profiles across four personalization dimensions. To facilitate profile supervision, we preprocess the ALOE training set by converting each user profile into a structured slot-value format, enabling fine-grained attribute tracking aligned with our reward design. The full construction procedure is detailed in Appendix \ref{app:slot_construction}.

\paragraph{Benchmarks} We evaluate on the ALOE benchmark \citep{wu2025aligning}, which provides multi-turn dialogues annotated with user profiles across diverse attributes for personalized dialogue evaluation.

We consider two settings: (1) \textbf{In-Format Generalization (Vanilla)}: Test users follow the same profile schema as training but contain unseen content, evaluating within-schema personalization. (2) \textbf{Cross-Format Generalization (Extended)}: Test users include both unseen attribute types and values, assessing the model's ability to infer profiles from dialogue without relying on fixed schemas.

We adopt the average alignment score (\textbf{AVG.}), normalized improvement ratio (\textbf{N-IR}) and normalized coefficient of determination (\textbf{N-$R^2$}). Details on metrics calculation and evaluation prompts are in Appendix \ref{app:metrics} and Appendix \ref{app:eval_prompt}, respectively. 

\paragraph{Baseline Methods} We compare RLPA against two major categories of baseline approaches for personalized alignment. Prompt-based Methods: (1) \textbf{Reminder} \citep{zhao2025llms}, (2) \textbf{Self-Critic} \citep{zhao2025llms}, (3) \textbf{Chain-of-Thought (CoT)} \citep{wei2022chain} and (4) \textbf{RAG} \citep{zhao2025llms}. Offline Optimization Methods: (5) \textbf{Supervised Finetuning (SFT)} \citep{ouyang2022training} and (6) \textbf{Direct Preference Optimization (DPO)} \citep{rafailov2023direct}. Please refer to Appendix \ref{app:baseline} for the detailed description of the baseline methods.

\paragraph{Implementation Details} We implement our RLPA training pipeline using the {OpenRLHF \citep{hu2024openrlhf} and vLLM \citep{kwon2023efficient} frameworks for scalable and stable reinforcement learning with LLMs. All experiments are conducted on 8 NVIDIA A100 80GB GPUs. For detailed hyper-parameter settings, please refer to Appendix \ref{app:implementation}.

\begin{table*}
\centering
\caption{Evaluation of personalized alignment performance on Vanilla and Extended ALOE settings. We report the average alignment score, as well as two auxiliary metrics: N-IR (normalized improvement rate) and N-$R^2$ (normalized coefficient of determination).}
\label{tab:alignment}
\resizebox{\linewidth}{!}{
\begin{tabular}{l c c c c | c c c }
\toprule
\multirow{2}{*}{\textbf{Model}} & \multirow{2}{*}{\textbf{Method}} 
& \multicolumn{3}{c|}{\textbf{Vanilla ALOE}} 
& \multicolumn{3}{c}{\textbf{Extended ALOE}} \\
\cmidrule(lr){3-5} \cmidrule(lr){6-8}
&
& \textbf{Align. Score (AVG.)} $\uparrow$ & \textbf{N-IR} $\uparrow$ & \textbf{N-}$R^2$ $\uparrow$
& \textbf{Align. Score (AVG.)} $\uparrow$ & \textbf{N-IR} $\uparrow$ & \textbf{N-}$R^2$ $\uparrow$ \\
\midrule
GPT-4o-mini & Self-Critic & 75.81 & 0.079 & 0.500 & 51.23 & 0.020 & 0.037 \\
GPT-4o & Self-Critic & 74.59 & 0.068 & 0.380 & 54.66 & 0.027 & 0.070 \\
Claude-3.5-Sonnet & Self-Critic & 69.19 & 0.105 & 0.792 & 35.62 & 0.054 & 0.266 \\
Claude-3.5-Haiku & Self-Critic & 59.19 & 0.075 & 0.461 & 34.93 & 0.046 & 0.200 \\
DeepSeek-V3 & Self-Critic & 50.81 & 0.055 & 0.268 & 39.45 & 0.033 & 0.106 \\
\midrule
\multirow{8}{*}{Qwen2.5-3B-Instruct} & Base & 7.97 & -0.020 & 0.047 & 1.78 & -0.042 & 0.083 \\
& Reminder & 17.03 & -0.001 & 0 & 7.26 & -0.034 & 0.084 \\
& Self-Critic & 52.43 & 0.056 & 0.243 & 7.26 & 0.023 & 0.046 \\
& CoT & 24.46 & 0 & 0 & 26.30 & -0.048 & 0.139 \\
& RAG(Top-5) & 28.65 & 0.001 & 0 & 11.64 & -0.030 & 0.042 \\
& SFT & 44.32 & 0.083 & 0.628 & 24.38 & 0.049 & 0.159 \\
& DPO & 45.27 & 0.070 & 0.389 & 27.26 & -0.021 & 0.037 \\
\cmidrule(lr){2-8}
\multirow{1}{*}{}& \textbf{RLPA (Ours)} & \textbf{73.38} & \textbf{0.090} & \textbf{0.855} & \textbf{52.74} & \textbf{0.100} & \textbf{0.498} \\
\bottomrule
\end{tabular}
}
\end{table*}

\subsection{Overall Results}

\begin{figure}[htbp]
    \centering
    \includegraphics[width=0.5\linewidth]{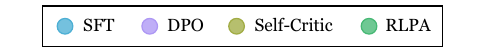}
    \vspace{50pt}
\end{figure}
\begin{figure*}[htbp]
    \vspace{-65pt}
    \centering
    \subfigure[Turn-wise alignment scores of SFT.]{\includegraphics[width=0.49\textwidth]{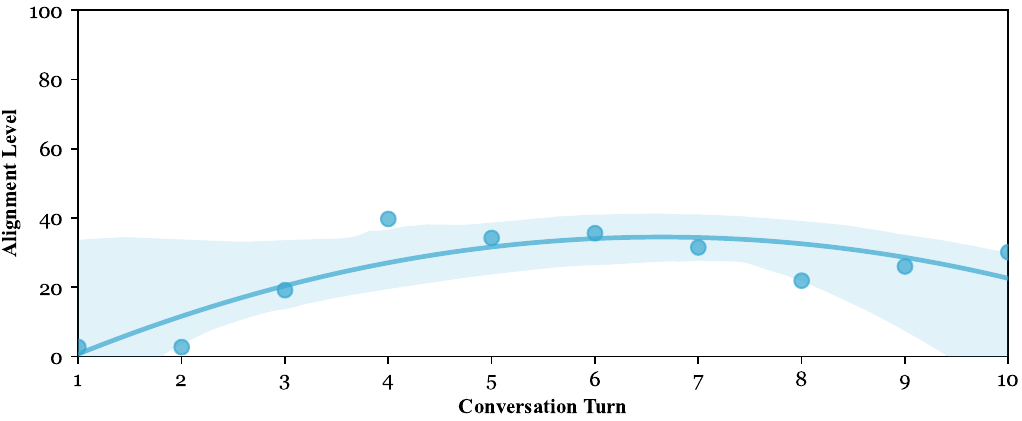}}
    \subfigure[Turn-wise alignment scores of DPO.]{\includegraphics[width=0.49\textwidth]{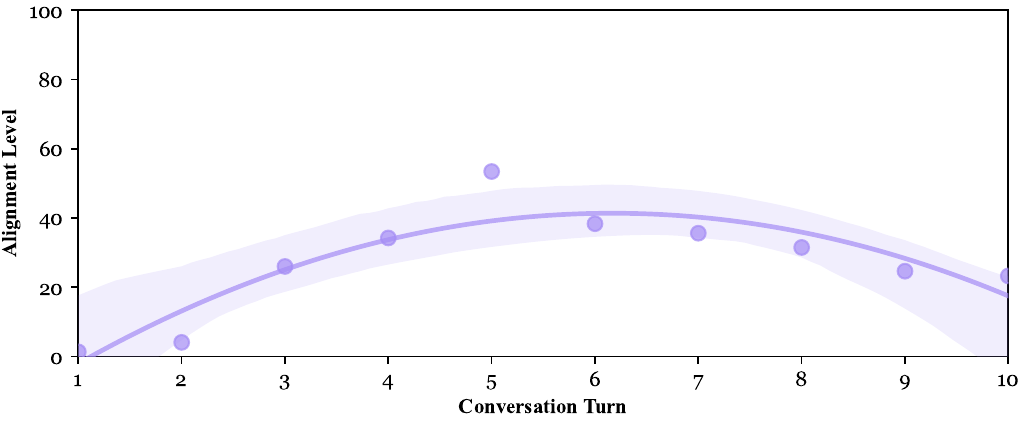}}\\
    \vspace{-10pt}
    \subfigure[Turn-wise alignment scores of Self-Critic.]{\includegraphics[width=0.49\textwidth]{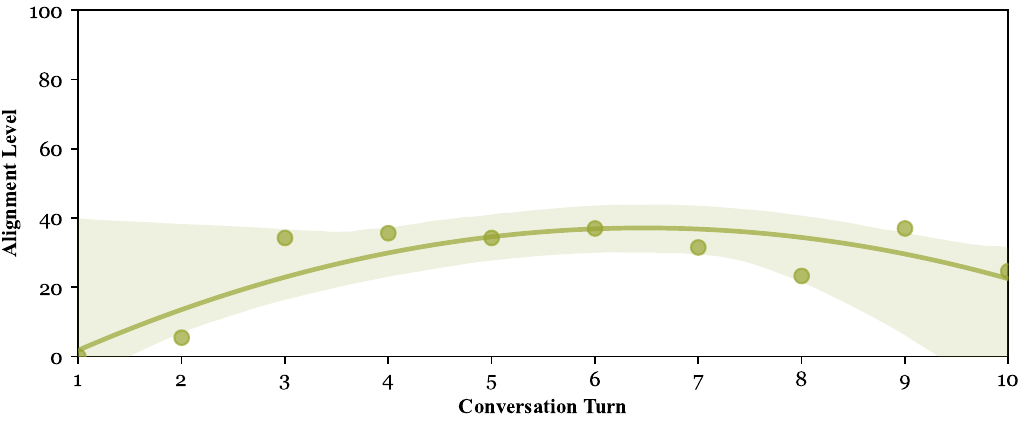}}
    \subfigure[Turn-wise alignment scores of RLPA.]{\includegraphics[width=0.49\textwidth]{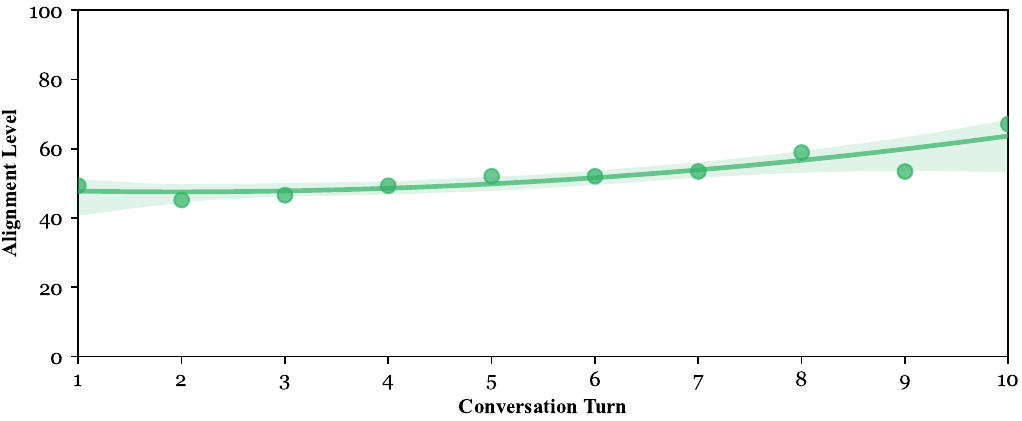}}
    \caption{Turn-wise alignment scores on the Extended ALOE benchmark across different personalization alignment methods, including (a) SFT, (b) DPO, (c) Self-Critic and (d) RLPA.}
    \label{fig:visual_all}
\end{figure*}

Table~\ref{tab:alignment} presents the comparison of personalized alignment performance across models on both Vanilla and Extended ALOE benchmarks. Our RLPA achieves the highest overall alignment scores in both settings,demonstrating strong personalized alignment under both familiar and schema-shifted scenarios. Compared to baselines built on the same backbone, RLPA offers consistent and substantial gains. In the Vanilla setting, it improves over SFT by +29.06 and over DPO by +28.11 in alignment score, and achieves the best N-R$^2$ (0.855) and N-IR (0.090) among all models, indicating not only higher personalization performance but also better profile–response consistency. In the Extended setting, RLPA outperforms SFT and DPO by over +28 points, validating its generalization ability to unseen profile formats. In comparison with closed-source LLMs, RLPA remains highly competitive. While GPT-4o and GPT-4o-mini achieve slightly higher raw alignment in the Vanilla setting, their N-R$^2$ scores are much lower (0.380 and 0.500), suggesting weaker response–profile coherence. On the Extended benchmark, RLPA leads in all three metrics, highlighting its robustness.

Figure~\ref{fig:visual_all} visualizes the turn-wise alignment scores on the Extended ALOE benchmark across four representative personalization methods. SFT and DPO  show early-stage improvement, peaking around turn 5, but their alignment scores degrade significantly in the later turns. This indicates that while these methods can initially adapt to user preferences, they struggle to maintain consistency across extended interactions. In contrast, RLPA demonstrates a consistently rising alignment trend, with stable progression across all 10 turns. This reflects its ability to continually refine the inferred profile and use it effectively for response generation, even as the dialogue evolves.

\section{Analysis and Discussions}

\begin{table*}
\centering
\caption{Ablation study on the impact of reward components.}
\label{tab:ablation}
\resizebox{\linewidth}{!}{
\begin{tabular}{l *{10}{c} c c c}
\toprule
\multirow{2.5}{*}{\textbf{Method}} & \multicolumn{11}{c}{\textbf{Alignment Level across kth Turn}} & \multicolumn{2}{c}{\textbf{Improvement Level}}\\
\cmidrule(lr){2-12}
& \textbf{1} & \textbf{2} & \textbf{3} & \textbf{4} & \textbf{5} & \textbf{6} & \textbf{7} & \textbf{8} & \textbf{9} & \textbf{10} & \textbf{AVG.} & \textbf{N-IR} & \textbf{N-}$R^2$ \\
\midrule
\rowcolor{gray!20} RLPA & \textbf{62.16} & \textbf{68.92} & \textbf{70.27} & \textbf{74.32} & \textbf{72.97} & \textbf{74.32} & \textbf{75.68} & \textbf{78.38} & \textbf{77.03} & \textbf{79.73} & \textbf{73.38} & \textbf{0.090} & \textbf{0.855} \\
\midrule
w/ PR & 41.89 & 33.78 & 50.00 & 51.35 & 51.35 & 44.59 & 47.30 & 47.30 & 50.00 & 37.84 & 45.54 & 0.015 & 0.019 \\
w/ RR & 58.92 & 63.92 & 60.27 & 67.03 & 60.27 & 72.97 & 64.32 & 73.78 & 71.08 & 69.32 & 66.19 & 0.088 & 0.524 \\
\bottomrule
\end{tabular}
}
\end{table*}

\subsection{Ablation Study}

To assess the individual contributions of the Profile Reward (PR) and Response Reward (RR), we conduct ablation experiments by disabling one reward component at a time during training. Table \ref{tab:ablation} reports the alignment results across dialogue turns. The reward progression curves for each training setup are provided in Appendix \ref{app:reward_curve}, indicating that training remains stable across all configurations, with smooth convergence patterns and no signs of reward collapse or instability.

We observe that removing either component leads to substantial performance drops. Using only the Profile Reward yields an average alignment score of 45.54, showing that while the model learns to infer user attributes, it struggles to reflect them fluently in response generation. In contrast, using only the Response Reward improves response-level personalization (AVG: 66.19) but lacks explicit supervision for profile construction, leading to weaker early-turn alignment (e.g., 58.92 at k=1) and instability in later turns. These results highlight the complementary roles of the two reward components: Profile Reward helps build accurate and structured user representations, while Response Reward ensures those representations are effectively utilized in generation.

\begin{figure*}
    \centering
    \subfigure[Alignment scores under a preference conflict setting, where the user profile is deliberately changed at turn 6.]{\includegraphics[width=0.49\textwidth]{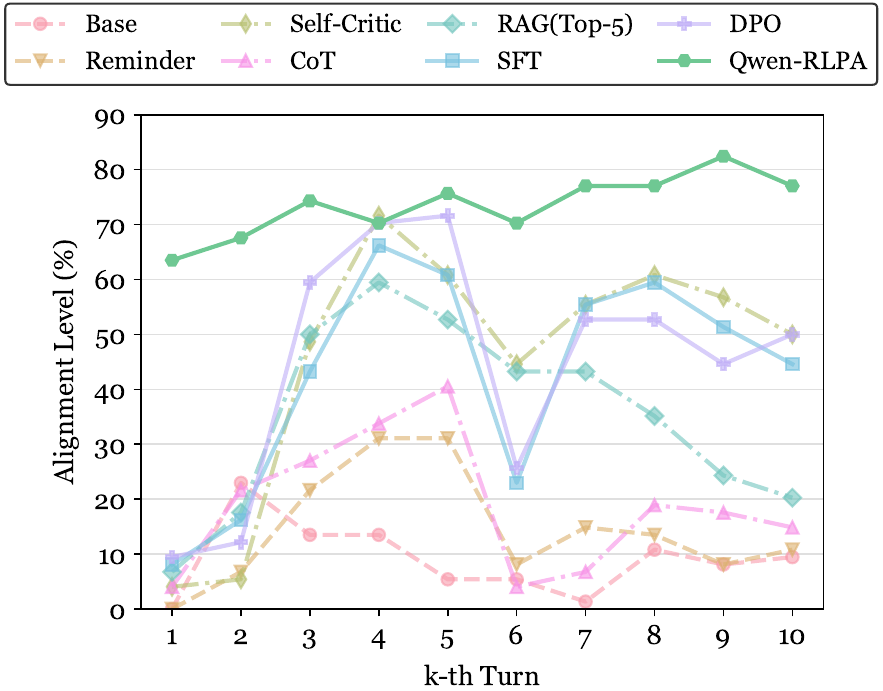}
    \label{subfig:conflict}}
    \subfigure[Long-term profile inference performance of Qwen-RLPA over 70 dialogue turns.]{\includegraphics[width=0.49\textwidth]{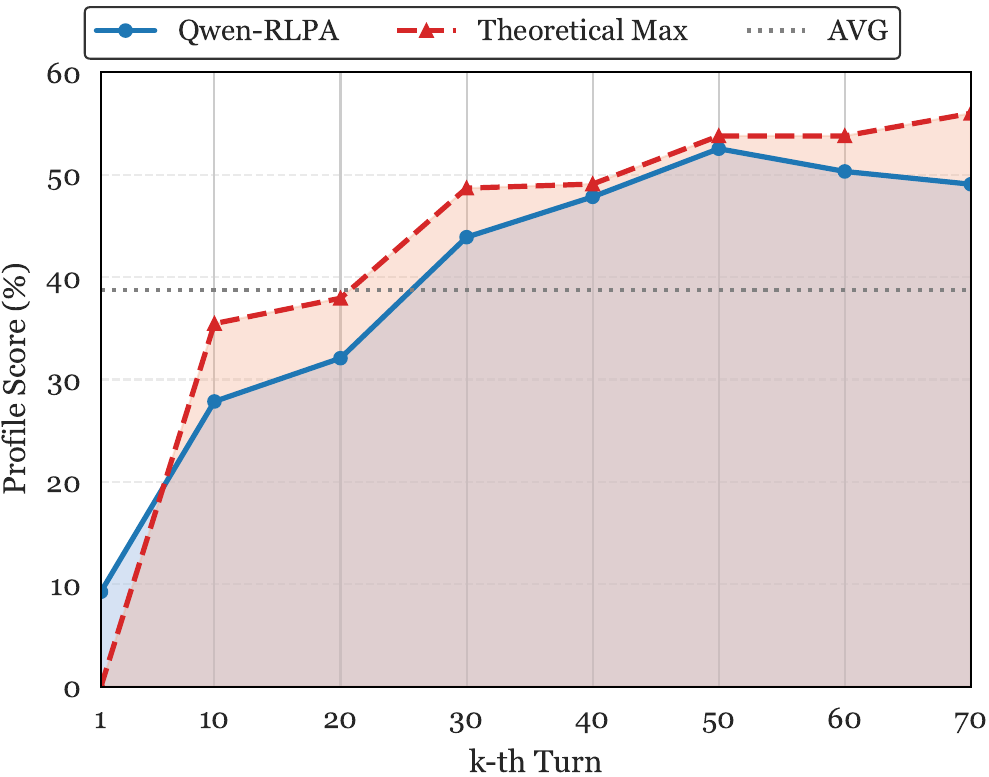}
    \label{subfig:long_turn}}
    \caption{Deeper analysis on Qwen-RLPA. (a) Performance under the preference conflict setting. (b) Performance of the long-term profile modeling.}
\end{figure*}

\subsection{Adaptation to Preference Conflict}

To evaluate Qwen-RLPA's adaptability to evolving user preferences, we introduce a preference shift at turn $k=6$, simulating real-world scenarios where user's preference may change mid-dialogue. As shown in Figure~\ref{subfig:conflict}, RLPA maintains high alignment despite the shift, with only a minor drop at turn 6 (70.27) and rapid recovery in later turns (e.g., 82.43 at $k=9$). This indicates that the model can detect profile changes and promptly adjust its behavior. In contrast, baselines such as DPO struggle under this setting—dropping sharply from 71.62 to 25.68 at turn 6—and show limited recovery, revealing poor responsiveness to dynamic user intent. These results highlight RLPA's strength in real-time profile adaptation, enabling it to revise and align with user preferences as they evolve.

\subsection{Stable Profile Modeling in Long-Term Interaction}

To evaluate the stability and reliability of user modeling over extended interactions, we conduct a long-term dialogue test lasting 70 turns, during which the user simulator consistently follows a fixed profile. We measure the Qwen-RLPA's profile accuracy at regular intervals by prompting it to generate an explicit profile summary and comparing it against the ground truth.

As shown in Figure \ref{subfig:long_turn}, the profile score increases steadily over time, from 9.26 at $k=1$ to 52.54 at $k=50$, and remains stable in the later stages (e.g., 50.32 at $k=60$, 49.07 at $k=70$). This result demonstrates that our Qwen-RLPA supports robust long-term profile inference, allowing it to accumulate user information over time and preserve it effectively throughout the interaction.

The AVG line averages profile scores at 8 checkpoints $(k = 1, 10, \ldots, 70)$, summarizing long-range tracking performance. The Theoretical Max represents the proportion of user attributes explicitly revealed (per ALOE annotation rules) up to each turn, reflecting the maximum recoverable profile. Qwen-RLPA’s near-convergence to this upper bound after turn 50 indicates its ability to capture and retain all available user cues with high fidelity.

\subsection{Comparison with Reasoning LLMs}
\begin{wrapfigure}{r}{0.5\textwidth}
  \vspace{-3.5mm}
  \centering
  \includegraphics[width=1.0\linewidth]{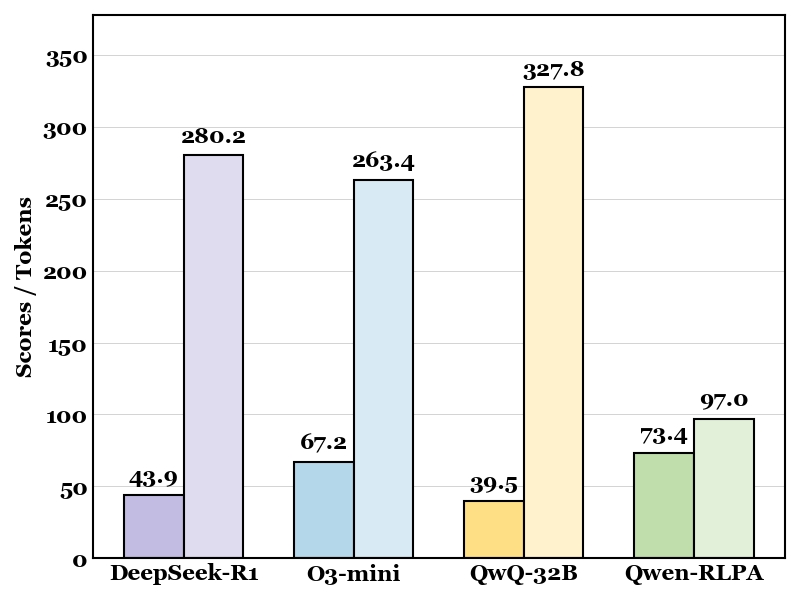}
  \vspace{-2mm}
  \caption{Comparison of response quality and reasoning efficiency across reasoning-centric models. Each pair of bars shows the average response alignment score (left) and the number of reasoning tokens used per turn (right).}
  \label{fig:reasoning_efficiency}
  \vspace{-4mm}
\end{wrapfigure}
We consider inferring user profiles as a domain-specific form of reasoning in personalized dialogue. Accordingly, we compare our Qwen-RLPA model against several reasoning-centric LLMs, including DeepSeek-R1 \citep{guo2025deepseek}, GPT-o3-mini \citep{openai2025o3blog}, and QwQ-32B \citep{qwen2025qwqblog}.

As shown in Figure \ref{fig:reasoning_efficiency}, Qwen-RLPA consistently achieves higher response scores while utilizing fewer reasoning tokens---those involved in inferring and maintaining user profiles. In contrast, models such as QwQ-32B generate over 300 tokens per turn yet fail to achieve comparable alignment, suggesting inefficiencies or misalignment in their reasoning processes. GPT-o3-mini follows a similar pattern, consuming more tokens but yielding lower response quality. These findings demonstrate that RLPA facilitates more focused, efficient, and profile-aware reasoning, surpassing general-purpose reasoning models in personalized dialogue tasks.

\section{Related Works}

Recent efforts have explored various strategies to personalize large language models for specific user needs, which can be divided into two categories.

On the one hand, prompt-based methods, including profile-augmented prompting \citep{wang2023chain, richardson2023integrating, pandey2024cos, li2024learning} and retrieval-augmented prompting \citep{zhang2024llm, li2024hello, salemi2024optimization, zhuang2024hydra,qiu2025measuring}, typically rely on static templates to inject user-specific information. While simple to implement, these methods offer only superficial personalization, are constrained by context length limitations \citep{liu2025survey}, and lack mechanisms for long-term memory or adaptive behavior.

On the other hand, offline optimization approaches, such as supervised fine-tuning (SFT) \citep{ouyang2022training, clarke2024peft, tan2024personalized, peng2024pocketllm} and direct preference optimization (DPO) \citep{rafailov2023direct, jang2023personalized, kirk2024prism, zollo2024personalllm, chen2025deeper}, aim to train models to produce profile-consistent responses. However, they require extensive labeled data, making them unsuitable for cold-start scenarios. Moreover, their static nature limits generalization across diverse users and hinders real-time adaptation \citep{xu2024dpo, chu2025sft}.

In summary, both prompting-based and offline methods struggle to achieve dynamic, long-term, and adaptive personalization, motivating the need for interactive learning frameworks such as ours.

\section{Conclusion}

In this work, we tackle the key challenge of enabling dynamic and effective personalization in LLMs, particularly under cold-start conditions and evolving user preferences. We formulate personalized alignment as a multi-turn Markov Decision Process and introduce RLPA, a reinforcement learning framework that empowers LLMs to infer, retain, and leverage user profiles through ongoing interaction. RLPA incorporates a dual-level reward scheme—combining profile-level and response-level feedback—which allows the model to adapt continuously to individual user needs. Our fine-tuned Qwen-RLPA model achieves substantial empirical gains, outperforming strong baselines across diverse personalization benchmarks. It also matches or exceeds the performance of leading proprietary systems, offering superior alignment quality and long-term consistency. Additionally, comparisons with recent reasoning LLMs highlight that profile-based reasoning, as facilitated by RLPA, represents a more efficient and contextually appropriate paradigm for personalized dialogue generation.

\section{Limitation and Future Works}

While our proposed RLPA framework demonstrates strong performance in dynamic personalized alignment, several limitations remain:

First, while RLPA supports cold-start adaptation, it currently assumes a single-user interaction thread. Extending the framework to multi-user, multi-session, or cross-domain personalization would better reflect real-world usage patterns.

Second, although we formalize personalization as a multi-turn MDP, the theoretical understanding of long-term alignment dynamics and convergence properties remains underexplored. Future research may investigate more principled frameworks for continual user modeling and policy generalization.

\section*{Acknowledgments}
We thank the anonymous reviewers for their comments and suggestions. This work was supported by the New Generation Artificial Intelligence-National Science and Technology Major Project 2023ZD0121100, the National Natural Science Foundation of China (NSFC) via grant 62441614 and 62176078, the Fundamental Research Funds for the Central Universities.

\bibliography{natbib}

@article{askell2021general,
  title={A general language assistant as a laboratory for alignment},
  author={Askell, Amanda and Bai, Yuntao and Chen, Anna and Drain, Dawn and Ganguli, Deep and Henighan, Tom and Jones, Andy and Joseph, Nicholas and Mann, Ben and DasSarma, Nova and others},
  journal={arXiv preprint arXiv:2112.00861},
  year={2021}
}

@article{bai2022constitutional,
  title={Constitutional ai: Harmlessness from ai feedback},
  author={Bai, Yuntao and Kadavath, Saurav and Kundu, Sandipan and Askell, Amanda and Kernion, Jackson and Jones, Andy and Chen, Anna and Goldie, Anna and Mirhoseini, Azalia and McKinnon, Cameron and others},
  journal={arXiv preprint arXiv:2212.08073},
  year={2022}
}

@article{zhao2023survey,
  title={A survey of large language models},
  author={Zhao, Wayne Xin and Zhou, Kun and Li, Junyi and Tang, Tianyi and Wang, Xiaolei and Hou, Yupeng and Min, Yingqian and Zhang, Beichen and Zhang, Junjie and Dong, Zican and others},
  journal={arXiv preprint arXiv:2303.18223},
  volume={1},
  number={2},
  year={2023}
}

@article{ouyang2022training,
  title={Training language models to follow instructions with human feedback},
  author={Ouyang, Long and Wu, Jeffrey and Jiang, Xu and Almeida, Diogo and Wainwright, Carroll and Mishkin, Pamela and Zhang, Chong and Agarwal, Sandhini and Slama, Katarina and Ray, Alex and others},
  journal={Advances in neural information processing systems},
  volume={35},
  pages={27730--27744},
  year={2022}
}

@article{ji2023ai,
  title={Ai alignment: A comprehensive survey},
  author={Ji, Jiaming and Qiu, Tianyi and Chen, Boyuan and Zhang, Borong and Lou, Hantao and Wang, Kaile and Duan, Yawen and He, Zhonghao and Zhou, Jiayi and Zhang, Zhaowei and others},
  journal={arXiv preprint arXiv:2310.19852},
  year={2023}
}

@article{shen2023large,
  title={Large language model alignment: A survey},
  author={Shen, Tianhao and Jin, Renren and Huang, Yufei and Liu, Chuang and Dong, Weilong and Guo, Zishan and Wu, Xinwei and Liu, Yan and Xiong, Deyi},
  journal={arXiv preprint arXiv:2309.15025},
  year={2023}
}

@inproceedings{sorensen2024position,
  title={Position: a roadmap to pluralistic alignment},
  author={Sorensen, Taylor and Moore, Jared and Fisher, Jillian and Gordon, Mitchell and Mireshghallah, Niloofar and Rytting, Christopher Michael and Ye, Andre and Jiang, Liwei and Lu, Ximing and Dziri, Nouha and others},
  booktitle={Proceedings of the 41st International Conference on Machine Learning},
  pages={46280--46302},
  year={2024}
}

@article{kirk2024benefits,
  title={The benefits, risks and bounds of personalizing the alignment of large language models to individuals},
  author={Kirk, Hannah Rose and Vidgen, Bertie and R{\"o}ttger, Paul and Hale, Scott A},
  journal={Nature Machine Intelligence},
  volume={6},
  number={4},
  pages={383--392},
  year={2024},
  publisher={Nature Publishing Group UK London}
}

@article{liu2025survey,
  title={A Survey of Personalized Large Language Models: Progress and Future Directions},
  author={Liu, Jiahong and Qiu, Zexuan and Li, Zhongyang and Dai, Quanyu and Zhu, Jieming and Hu, Minda and Yang, Menglin and King, Irwin},
  journal={arXiv preprint arXiv:2502.11528},
  year={2025}
}

@article{zhang2024personalization,
  title={Personalization of large language models: A survey},
  author={Zhang, Zhehao and Rossi, Ryan A and Kveton, Branislav and Shao, Yijia and Yang, Diyi and Zamani, Hamed and Dernoncourt, Franck and Barrow, Joe and Yu, Tong and Kim, Sungchul and others},
  journal={arXiv preprint arXiv:2411.00027},
  year={2024}
}

@inproceedings{tseng2024two,
  title={Two Tales of Persona in LLMs: A Survey of Role-Playing and Personalization},
  author={Tseng, Yu-Min and Huang, Yu-Chao and Hsiao, Teng-Yun and Chen, Wei-Lin and Huang, Chao-Wei and Meng, Yu and Chen, Yun-Nung},
  booktitle={Findings of the Association for Computational Linguistics: EMNLP 2024},
  pages={16612--16631},
  year={2024}
}

@article{wu2024personalized,
  title={Personalized Multimodal Large Language Models: A Survey},
  author={Wu, Junda and Lyu, Hanjia and Xia, Yu and Zhang, Zhehao and Barrow, Joe and Kumar, Ishita and Mirtaheri, Mehrnoosh and Chen, Hongjie and Rossi, Ryan A and Dernoncourt, Franck and others},
  journal={arXiv preprint arXiv:2412.02142},
  year={2024}
}

@inproceedings{salemi2024lamp,
  title={LaMP: When Large Language Models Meet Personalization},
  author={Salemi, Alireza and Mysore, Sheshera and Bendersky, Michael and Zamani, Hamed},
  booktitle={Proceedings of the 62nd Annual Meeting of the Association for Computational Linguistics (Volume 1: Long Papers)},
  pages={7370--7392},
  year={2024}
}

@inproceedings{jiang2024peek,
  title={A Peek into Token Bias: Large Language Models Are Not Yet Genuine Reasoners},
  author={Jiang, Bowen and Xie, Yangxinyu and Hao, Zhuoqun and Wang, Xiaomeng and Mallick, Tanwi and Su, Weijie and Taylor, Camillo and Roth, Dan},
  booktitle={Proceedings of the 2024 Conference on Empirical Methods in Natural Language Processing},
  pages={4722--4756},
  year={2024}
}

@article{jiang2025know,
  title={Know Me, Respond to Me: Benchmarking LLMs for Dynamic User Profiling and Personalized Responses at Scale},
  author={Jiang, Bowen and Hao, Zhuoqun and Cho, Young-Min and Li, Bryan and Yuan, Yuan and Chen, Sihao and Ungar, Lyle and Taylor, Camillo J and Roth, Dan},
  journal={arXiv preprint arXiv:2504.14225},
  year={2025}
}

@inproceedings{shi2024wildfeedback,
  title={WildFeedback: Aligning LLMs With In-situ User Interactions And Feedback},
  author={Shi, Taiwei and Wang, Zhuoer and Yang, Longqi and Lin, Ying-Chun and He, Zexue and Wan, Mengting and Zhou, Pei and Jauhar, Sujay Kumar and Xu, Xiaofeng and Song, Xia and others},
  booktitle={NeurIPS 2024 Workshop on Behavioral Machine Learning},
  year={2024}
}

@article{zhao2025llms,
  title={Do LLMs Recognize Your Preferences? Evaluating Personalized Preference Following in LLMs},
  author={Zhao, Siyan and Hong, Mingyi and Liu, Yang and Hazarika, Devamanyu and Lin, Kaixiang},
  journal={arXiv preprint arXiv:2502.09597},
  year={2025}
}

@article{xie2025survey,
  title={A Survey on Personalized and Pluralistic Preference Alignment in Large Language Models},
  author={Xie, Zhouhang and Wu, Junda and Shen, Yiran and Xia, Yu and Li, Xintong and Chang, Aaron and Rossi, Ryan and Kumar, Sachin and Majumder, Bodhisattwa Prasad and Shang, Jingbo and others},
  journal={arXiv preprint arXiv:2504.07070},
  year={2025}
}

@article{wang2023chain,
  title={Chain-of-thought prompting for responding to in-depth dialogue questions with LLM},
  author={Wang, Hongru and Wang, Rui and Mi, Fei and Wang, Zezhong and Xu, Ruifeng and Wong, Kam-Fai},
  journal={arXiv preprint arXiv:2305.11792},
  year={2023}
}

@article{richardson2023integrating,
  title={Integrating summarization and retrieval for enhanced personalization via large language models},
  author={Richardson, Chris and Zhang, Yao and Gillespie, Kellen and Kar, Sudipta and Singh, Arshdeep and Raeesy, Zeynab and Khan, Omar Zia and Sethy, Abhinav},
  journal={arXiv preprint arXiv:2310.20081},
  year={2023}
}

@inproceedings{pandey2024cos,
  title={CoS: Enhancing Personalization and Mitigating Bias with Context Steering},
  author={Pandey, Sashrika and He, Zhiyang and Schrum, Mariah and Dragan, Anca},
  booktitle={Interpretable AI: Past, Present and Future},
  year={2024}
}

@inproceedings{li2024learning,
  title={Learning to rewrite prompts for personalized text generation},
  author={Li, Cheng and Zhang, Mingyang and Mei, Qiaozhu and Kong, Weize and Bendersky, Michael},
  booktitle={Proceedings of the ACM Web Conference 2024},
  pages={3367--3378},
  year={2024}
}

@inproceedings{zhang2024llm,
  title={LLM-based Medical Assistant Personalization with Short-and Long-Term Memory Coordination},
  author={Zhang, Kai and Kang, Yangyang and Zhao, Fubang and Liu, Xiaozhong},
  booktitle={Proceedings of the 2024 Conference of the North American Chapter of the Association for Computational Linguistics: Human Language Technologies (Volume 1: Long Papers)},
  pages={2386--2398},
  year={2024}
}

@article{li2024hello,
  title={Hello again! llm-powered personalized agent for long-term dialogue},
  author={Li, Hao and Yang, Chenghao and Zhang, An and Deng, Yang and Wang, Xiang and Chua, Tat-Seng},
  journal={arXiv preprint arXiv:2406.05925},
  year={2024}
}

@inproceedings{salemi2024optimization,
  title={Optimization methods for personalizing large language models through retrieval augmentation},
  author={Salemi, Alireza and Kallumadi, Surya and Zamani, Hamed},
  booktitle={Proceedings of the 47th International ACM SIGIR Conference on Research and Development in Information Retrieval},
  pages={752--762},
  year={2024}
}

@inproceedings{zhuang2024hydra,
  title={HYDRA: Model Factorization Framework for Black-Box LLM Personalization},
  author={Zhuang, Yuchen and Sun, Haotian and Yu, Yue and Qiang, Rushi and Wang, Qifan and Zhang, Chao and Dai, Bo},
  booktitle={The Thirty-eighth Annual Conference on Neural Information Processing Systems},
  year={2024}
}

@article{rafailov2023direct,
  title={Direct preference optimization: Your language model is secretly a reward model},
  author={Rafailov, Rafael and Sharma, Archit and Mitchell, Eric and Manning, Christopher D and Ermon, Stefano and Finn, Chelsea},
  journal={Advances in Neural Information Processing Systems},
  volume={36},
  pages={53728--53741},
  year={2023}
}

@article{chu2025sft,
  title={Sft memorizes, rl generalizes: A comparative study of foundation model post-training},
  author={Chu, Tianzhe and Zhai, Yuexiang and Yang, Jihan and Tong, Shengbang and Xie, Saining and Schuurmans, Dale and Le, Quoc V and Levine, Sergey and Ma, Yi},
  journal={arXiv preprint arXiv:2501.17161},
  year={2025}
}

@inproceedings{lin2024limited,
  title={On the Limited Generalization Capability of the Implicit Reward Model Induced by Direct Preference Optimization},
  author={Lin, Yong and Seto, Skyler and Ter Hoeve, Maartje and Metcalf, Katherine and Theobald, Barry-John and Wang, Xuan and Zhang, Yizhe and Huang, Chen and Zhang, Tong},
  booktitle={Findings of the Association for Computational Linguistics: EMNLP 2024},
  pages={16015--16026},
  year={2024}
}

@inproceedings{xu2024dpo,
  title={Is DPO Superior to PPO for LLM Alignment? A Comprehensive Study},
  author={Xu, Shusheng and Fu, Wei and Gao, Jiaxuan and Ye, Wenjie and Liu, Weilin and Mei, Zhiyu and Wang, Guangju and Yu, Chao and Wu, Yi},
  booktitle={International Conference on Machine Learning},
  pages={54983--54998},
  year={2024},
  organization={PMLR}
}

@article{clarke2024peft,
  title={PEFT-U: Parameter-Efficient Fine-Tuning for User Personalization},
  author={Clarke, Christopher and Heng, Yuzhao and Tang, Lingjia and Mars, Jason},
  journal={arXiv preprint arXiv:2407.18078},
  year={2024}
}

@inproceedings{tan2024personalized,
  title={Personalized Pieces: Efficient Personalized Large Language Models through Collaborative Efforts},
  author={Tan, Zhaoxuan and Liu, Zheyuan and Jiang, Meng},
  booktitle={Proceedings of the 2024 Conference on Empirical Methods in Natural Language Processing},
  pages={6459--6475},
  year={2024}
}

@inproceedings{peng2024pocketllm,
  title={PocketLLM: Enabling On-Device Fine-Tuning for Personalized LLMs},
  author={Peng, Dan and Fu, Zhihui and Wang, Jun},
  booktitle={Proceedings of the Fifth Workshop on Privacy in Natural Language Processing},
  pages={91--96},
  year={2024}
}

@article{kirk2024prism,
  title={The prism alignment project: What participatory, representative and individualised human feedback reveals about the subjective and multicultural alignment of large language models},
  author={Kirk, Hannah Rose and Whitefield, Alexander and R{\"o}ttger, Paul and Bean, Andrew and Margatina, Katerina and Ciro, Juan and Mosquera, Rafael and Bartolo, Max and Williams, Adina and He, He and others},
  journal={arXiv preprint arXiv:2404.16019},
  year={2024}
}

@inproceedings{zollo2024personalllm,
  title={PersonalLLM: Tailoring LLMs to Individual Preferences},
  author={Zollo, Thomas P and Siah, Andrew Wei Tung and Ye, Naimeng and Li, Ang and Namkoong, Hongseok},
  booktitle={The Thirteenth International Conference on Learning Representations},
  year={2024}
}

@inproceedings{jang2023personalized,
  title={Personalized Soups: Personalized Large Language Model Alignment via Post-hoc Parameter Merging},
  author={Jang, Joel and Kim, Seungone and Lin, Bill Yuchen and Wang, Yizhong and Hessel, Jack and Zettlemoyer, Luke and Hajishirzi, Hannaneh and Choi, Yejin and Ammanabrolu, Prithviraj},
  booktitle={Adaptive Foundation Models: Evolving AI for Personalized and Efficient Learning},
  year={2023}
}

@article{chen2025deeper,
  title={DEEPER Insight into Your User: Directed Persona Refinement for Dynamic Persona Modeling},
  author={Chen, Aili and Du, Chengyu and Chen, Jiangjie and Xu, Jinghan and Zhang, Yikai and Yuan, Siyu and Chen, Zulong and Li, Liangyue and Xiao, Yanghua},
  journal={arXiv preprint arXiv:2502.11078},
  year={2025}
}

@article{yang2024qwen2,
  title={Qwen2. 5 Technical Report},
  author={Yang, An and Yang, Baosong and Zhang, Beichen and Hui, Binyuan and Zheng, Bo and Yu, Bowen and Li, Chengyuan and Liu, Dayiheng and Huang, Fei and Wei, Haoran and others},
  journal={arXiv preprint arXiv:2412.15115},
  year={2024}
}

@article{openai2025o3blog,
  author = {OpenAI},
  title = {OpenAI o3-mini System Card},
  journal = {OpenAI's Blog},
  year = {2025},
  url = {https://openai.com/index/o3-mini-system-card}
}

@article{guo2025deepseek,
  title={Deepseek-r1: Incentivizing reasoning capability in llms via reinforcement learning},
  author={Guo, Daya and Yang, Dejian and Zhang, Haowei and Song, Junxiao and Zhang, Ruoyu and Xu, Runxin and Zhu, Qihao and Ma, Shirong and Wang, Peiyi and Bi, Xiao and others},
  journal={arXiv preprint arXiv:2501.12948},
  year={2025}
}

@article{hu2024openrlhf,
  title={OpenRLHF: An Easy-to-use, Scalable and High-performance RLHF Framework},
  author={Jian Hu and Xibin Wu and Zilin Zhu and Xianyu and Weixun Wang and Dehao Zhang and Yu Cao},
  journal={arXiv preprint arXiv:2405.11143},
  year={2024}
}

@inproceedings{kwon2023efficient,
  title={Efficient Memory Management for Large Language Model Serving with PagedAttention},
  author={Woosuk Kwon and Zhuohan Li and Siyuan Zhuang and Ying Sheng and Lianmin Zheng and Cody Hao Yu and Joseph E. Gonzalez and Hao Zhang and Ion Stoica},
  booktitle={Proceedings of the ACM SIGOPS 29th Symposium on Operating Systems Principles},
  year={2023}
}

@inproceedings{wu2025aligning,
  title={Aligning LLMs with Individual Preferences via Interaction},
  author={Wu, Shujin and Fung, Yi R and Qian, Cheng and Kim, Jeonghwan and Hakkani-Tur, Dilek and Ji, Heng},
  booktitle={Proceedings of the 31st International Conference on Computational Linguistics},
  pages={7648--7662},
  year={2025}
}

@article{wei2022chain,
  title={Chain-of-thought prompting elicits reasoning in large language models},
  author={Wei, Jason and Wang, Xuezhi and Schuurmans, Dale and Bosma, Maarten and Xia, Fei and Chi, Ed and Le, Quoc V and Zhou, Denny and others},
  journal={Advances in neural information processing systems},
  volume={35},
  pages={24824--24837},
  year={2022}
}

@article{schulman2017proximal,
  title={Proximal policy optimization algorithms},
  author={Schulman, John and Wolski, Filip and Dhariwal, Prafulla and Radford, Alec and Klimov, Oleg},
  journal={arXiv preprint arXiv:1707.06347},
  year={2017}
}

@article{schulman2015high,
  title={High-dimensional continuous control using generalized advantage estimation},
  author={Schulman, John and Moritz, Philipp and Levine, Sergey and Jordan, Michael and Abbeel, Pieter},
  journal={arXiv preprint arXiv:1506.02438},
  year={2015}
}

@article{qwen2025qwqblog,
  author = {Team Qwen},
  title = {QwQ-32B: Embracing the Power of Reinforcement Learning},
  journal = {Qwen's Blog},
  year = {2025},
  url = {https://qwenlm.github.io/blog/qwq-32b}
}

@article{brown2020language,
  title={Language models are few-shot learners},
  author={Brown, Tom and Mann, Benjamin and Ryder, Nick and Subbiah, Melanie and Kaplan, Jared D and Dhariwal, Prafulla and Neelakantan, Arvind and Shyam, Pranav and Sastry, Girish and Askell, Amanda and others},
  journal={Advances in neural information processing systems},
  volume={33},
  pages={1877--1901},
  year={2020}
}

@inproceedings{kojima2022large,
  title={Large Language Models are Zero-Shot Reasoners},
  author={Kojima, Takeshi and Gu, Shixiang Shane and Reid, Machel and Matsuo, Yutaka and Iwasawa, Yusuke},
  booktitle={Advances in Neural Information Processing Systems},
  year={2022}
}

@article{zhao2025beware,
  title={Beware of your po! measuring and mitigating ai safety risks in role-play fine-tuning of llms},
  author={Zhao, Weixiang and Hu, Yulin and Deng, Yang and Guo, Jiahe and Sui, Xingyu and Han, Xinyang and Zhang, An and Zhao, Yanyan and Qin, Bing and Chua, Tat-Seng and others},
  journal={arXiv preprint arXiv:2502.20968},
  year={2025}
}

@article{zhang2018personalizing,
  title={Personalizing dialogue agents: I have a dog, do you have pets too?},
  author={Zhang, Saizheng and Dinan, Emily and Urbanek, Jack and Szlam, Arthur and Kiela, Douwe and Weston, Jason},
  journal={arXiv preprint arXiv:1801.07243},
  year={2018}
}

@article{zheng2023judging,
  title={Judging llm-as-a-judge with mt-bench and chatbot arena},
  author={Zheng, Lianmin and Chiang, Wei-Lin and Sheng, Ying and Zhuang, Siyuan and Wu, Zhanghao and Zhuang, Yonghao and Lin, Zi and Li, Zhuohan and Li, Dacheng and Xing, Eric and others},
  journal={Advances in Neural Information Processing Systems},
  volume={36},
  pages={46595--46623},
  year={2023}
}

@article{gao2021simcse,
  title={Simcse: Simple contrastive learning of sentence embeddings},
  author={Gao, Tianyu and Yao, Xingcheng and Chen, Danqi},
  journal={arXiv preprint arXiv:2104.08821},
  year={2021}
}

@article{qiu2025measuring,
  title={Measuring What Makes You Unique: Difference-Aware User Modeling for Enhancing LLM Personalization},
  author={Qiu, Yilun and Zhao, Xiaoyan and Zhang, Yang and Bai, Yimeng and Wang, Wenjie and Cheng, Hong and Feng, Fuli and Chua, Tat-Seng},
  journal={arXiv preprint arXiv:2503.02450},
  year={2025}
}

@inproceedings{DBLP:conf/nips/ZhengWZN0C24,
  author       = {Jingnan Zheng and
                  Han Wang and
                  An Zhang and
                  Tai D. Nguyen and
                  Jun Sun and
                  Tat{-}Seng Chua},
  title        = {ALI-Agent: Assessing LLMs' Alignment with Human Values via Agent-based
                  Evaluation},
  booktitle    = {Advances in Neural Information Processing Systems 38: Annual Conference
                  on Neural Information Processing Systems 2024, NeurIPS 2024, Vancouver,
                  BC, Canada, December 10 - 15, 2024},
  year         = {2024},
}
\bibliographystyle{unsrtnat}

\newpage
\section*{NeurIPS Paper Checklist}

\begin{enumerate}

\item {\bf Claims}
    \item[] Question: Do the main claims made in the abstract and introduction accurately reflect the paper's contributions and scope?
    \item[] Answer: \answerYes{} 
    \item[] Justification: The main claims are described in Abstract and Section 1.
    \item[] Guidelines:
    \begin{itemize}
        \item The answer NA means that the abstract and introduction do not include the claims made in the paper.
        \item The abstract and/or introduction should clearly state the claims made, including the contributions made in the paper and important assumptions and limitations. A No or NA answer to this question will not be perceived well by the reviewers. 
        \item The claims made should match theoretical and experimental results, and reflect how much the results can be expected to generalize to other settings. 
        \item It is fine to include aspirational goals as motivation as long as it is clear that these goals are not attained by the paper. 
    \end{itemize}

\item {\bf Limitations}
    \item[] Question: Does the paper discuss the limitations of the work performed by the authors?
    \item[] Answer: \answerYes{} 
    \item[] Justification: We discuss the limitations of the work in Appendix A.
    \item[] Guidelines:
    \begin{itemize}
        \item The answer NA means that the paper has no limitation while the answer No means that the paper has limitations, but those are not discussed in the paper. 
        \item The authors are encouraged to create a separate "Limitations" section in their paper.
        \item The paper should point out any strong assumptions and how robust the results are to violations of these assumptions (e.g., independence assumptions, noiseless settings, model well-specification, asymptotic approximations only holding locally). The authors should reflect on how these assumptions might be violated in practice and what the implications would be.
        \item The authors should reflect on the scope of the claims made, e.g., if the approach was only tested on a few datasets or with a few runs. In general, empirical results often depend on implicit assumptions, which should be articulated.
        \item The authors should reflect on the factors that influence the performance of the approach. For example, a facial recognition algorithm may perform poorly when image resolution is low or images are taken in low lighting. Or a speech-to-text system might not be used reliably to provide closed captions for online lectures because it fails to handle technical jargon.
        \item The authors should discuss the computational efficiency of the proposed algorithms and how they scale with dataset size.
        \item If applicable, the authors should discuss possible limitations of their approach to address problems of privacy and fairness.
        \item While the authors might fear that complete honesty about limitations might be used by reviewers as grounds for rejection, a worse outcome might be that reviewers discover limitations that aren't acknowledged in the paper. The authors should use their best judgment and recognize that individual actions in favor of transparency play an important role in developing norms that preserve the integrity of the community. Reviewers will be specifically instructed to not penalize honesty concerning limitations.
    \end{itemize}

\item {\bf Theory assumptions and proofs}
    \item[] Question: For each theoretical result, does the paper provide the full set of assumptions and a complete (and correct) proof?
    \item[] Answer: \answerNA{} 
    \item[] Justification: The paper is empirical study and does not include theoretical results.
    \item[] Guidelines:
    \begin{itemize}
        \item The answer NA means that the paper does not include theoretical results. 
        \item All the theorems, formulas, and proofs in the paper should be numbered and cross-referenced.
        \item All assumptions should be clearly stated or referenced in the statement of any theorems.
        \item The proofs can either appear in the main paper or the supplemental material, but if they appear in the supplemental material, the authors are encouraged to provide a short proof sketch to provide intuition. 
        \item Inversely, any informal proof provided in the core of the paper should be complemented by formal proofs provided in appendix or supplemental material.
        \item Theorems and Lemmas that the proof relies upon should be properly referenced. 
    \end{itemize}

    \item {\bf Experimental result reproducibility}
    \item[] Question: Does the paper fully disclose all the information needed to reproduce the main experimental results of the paper to the extent that it affects the main claims and/or conclusions of the paper (regardless of whether the code and data are provided or not)?
    \item[] Answer: \answerYes{} 
    \item[] Justification: Implementation Details can be found in Section 4.1 and Appendix F.
    \item[] Guidelines:
    \begin{itemize}
        \item The answer NA means that the paper does not include experiments.
        \item If the paper includes experiments, a No answer to this question will not be perceived well by the reviewers: Making the paper reproducible is important, regardless of whether the code and data are provided or not.
        \item If the contribution is a dataset and/or model, the authors should describe the steps taken to make their results reproducible or verifiable. 
        \item Depending on the contribution, reproducibility can be accomplished in various ways. For example, if the contribution is a novel architecture, describing the architecture fully might suffice, or if the contribution is a specific model and empirical evaluation, it may be necessary to either make it possible for others to replicate the model with the same dataset, or provide access to the model. In general. releasing code and data is often one good way to accomplish this, but reproducibility can also be provided via detailed instructions for how to replicate the results, access to a hosted model (e.g., in the case of a large language model), releasing of a model checkpoint, or other means that are appropriate to the research performed.
        \item While NeurIPS does not require releasing code, the conference does require all submissions to provide some reasonable avenue for reproducibility, which may depend on the nature of the contribution. For example
        \begin{enumerate}
            \item If the contribution is primarily a new algorithm, the paper should make it clear how to reproduce that algorithm.
            \item If the contribution is primarily a new model architecture, the paper should describe the architecture clearly and fully.
            \item If the contribution is a new model (e.g., a large language model), then there should either be a way to access this model for reproducing the results or a way to reproduce the model (e.g., with an open-source dataset or instructions for how to construct the dataset).
            \item We recognize that reproducibility may be tricky in some cases, in which case authors are welcome to describe the particular way they provide for reproducibility. In the case of closed-source models, it may be that access to the model is limited in some way (e.g., to registered users), but it should be possible for other researchers to have some path to reproducing or verifying the results.
        \end{enumerate}
    \end{itemize}

\item {\bf Open access to data and code}
    \item[] Question: Does the paper provide open access to the data and code, with sufficient instructions to faithfully reproduce the main experimental results, as described in supplemental material?
    \item[] Answer: \answerYes{} 
    \item[] Justification: Our data and codes could be found in supplementary files.
    \item[] Guidelines:
    \begin{itemize}
        \item The answer NA means that paper does not include experiments requiring code.
        \item Please see the NeurIPS code and data submission guidelines (\url{https://nips.cc/public/guides/CodeSubmissionPolicy}) for more details.
        \item While we encourage the release of code and data, we understand that this might not be possible, so “No” is an acceptable answer. Papers cannot be rejected simply for not including code, unless this is central to the contribution (e.g., for a new open-source benchmark).
        \item The instructions should contain the exact command and environment needed to run to reproduce the results. See the NeurIPS code and data submission guidelines (\url{https://nips.cc/public/guides/CodeSubmissionPolicy}) for more details.
        \item The authors should provide instructions on data access and preparation, including how to access the raw data, preprocessed data, intermediate data, and generated data, etc.
        \item The authors should provide scripts to reproduce all experimental results for the new proposed method and baselines. If only a subset of experiments are reproducible, they should state which ones are omitted from the script and why.
        \item At submission time, to preserve anonymity, the authors should release anonymized versions (if applicable).
        \item Providing as much information as possible in supplemental material (appended to the paper) is recommended, but including URLs to data and code is permitted.
    \end{itemize}

\item {\bf Experimental setting/details}
    \item[] Question: Does the paper specify all the training and test details (e.g., data splits, hyperparameters, how they were chosen, type of optimizer, etc.) necessary to understand the results?
    \item[] Answer: \answerYes{} 
    \item[] Justification: Implementation Details can be found in Section 4.1 and Appendix F.
    \item[] Guidelines: 
    \begin{itemize}
        \item The answer NA means that the paper does not include experiments.
        \item The experimental setting should be presented in the core of the paper to a level of detail that is necessary to appreciate the results and make sense of them.
        \item The full details can be provided either with the code, in appendix, or as supplemental material.
    \end{itemize}

\item {\bf Experiment statistical significance}
    \item[] Question: Does the paper report error bars suitably and correctly defined or other appropriate information about the statistical significance of the experiments?
    \item[] Answer: \answerYes{} 
    \item[] Justification: Statistical significance of the experiments can be found in Appendix F.
    \item[] Guidelines:
    \begin{itemize}
        \item The answer NA means that the paper does not include experiments.
        \item The authors should answer "Yes" if the results are accompanied by error bars, confidence intervals, or statistical significance tests, at least for the experiments that support the main claims of the paper.
        \item The factors of variability that the error bars are capturing should be clearly stated (for example, train/test split, initialization, random drawing of some parameter, or overall run with given experimental conditions).
        \item The method for calculating the error bars should be explained (closed form formula, call to a library function, bootstrap, etc.)
        \item The assumptions made should be given (e.g., Normally distributed errors).
        \item It should be clear whether the error bar is the standard deviation or the standard error of the mean.
        \item It is OK to report 1-sigma error bars, but one should state it. The authors should preferably report a 2-sigma error bar than state that they have a 96\% CI, if the hypothesis of Normality of errors is not verified.
        \item For asymmetric distributions, the authors should be careful not to show in tables or figures symmetric error bars that would yield results that are out of range (e.g. negative error rates).
        \item If error bars are reported in tables or plots, The authors should explain in the text how they were calculated and reference the corresponding figures or tables in the text.
    \end{itemize}

\item {\bf Experiments compute resources}
    \item[] Question: For each experiment, does the paper provide sufficient information on the computer resources (type of compute workers, memory, time of execution) needed to reproduce the experiments?
    \item[] Answer: \answerYes{} 
    \item[] Justification: Experiments compute resources can be found in Appendix F.
    \item[] Guidelines:
    \begin{itemize}
        \item The answer NA means that the paper does not include experiments.
        \item The paper should indicate the type of compute workers CPU or GPU, internal cluster, or cloud provider, including relevant memory and storage.
        \item The paper should provide the amount of compute required for each of the individual experimental runs as well as estimate the total compute. 
        \item The paper should disclose whether the full research project required more compute than the experiments reported in the paper (e.g., preliminary or failed experiments that didn't make it into the paper). 
    \end{itemize}
    
\item {\bf Code of ethics}
    \item[] Question: Does the research conducted in the paper conform, in every respect, with the NeurIPS Code of Ethics \url{https://neurips.cc/public/EthicsGuidelines}?
    \item[] Answer: \answerYes{} 
    \item[] Justification: The research conducted in the paper conform, in every respect, with the NeurIPS Code of Ethics.
    \item[] Guidelines:
    \begin{itemize}
        \item The answer NA means that the authors have not reviewed the NeurIPS Code of Ethics.
        \item If the authors answer No, they should explain the special circumstances that require a deviation from the Code of Ethics.
        \item The authors should make sure to preserve anonymity (e.g., if there is a special consideration due to laws or regulations in their jurisdiction).
    \end{itemize}

\item {\bf Broader impacts}
    \item[] Question: Does the paper discuss both potential positive societal impacts and negative societal impacts of the work performed?
    \item[] Answer: \answerYes{} 
    \item[] Justification: Broader impacts can be found in Appendix B.
    \item[] Guidelines:
    \begin{itemize}
        \item The answer NA means that there is no societal impact of the work performed.
        \item If the authors answer NA or No, they should explain why their work has no societal impact or why the paper does not address societal impact.
        \item Examples of negative societal impacts include potential malicious or unintended uses (e.g., disinformation, generating fake profiles, surveillance), fairness considerations (e.g., deployment of technologies that could make decisions that unfairly impact specific groups), privacy considerations, and security considerations.
        \item The conference expects that many papers will be foundational research and not tied to particular applications, let alone deployments. However, if there is a direct path to any negative applications, the authors should point it out. For example, it is legitimate to point out that an improvement in the quality of generative models could be used to generate deepfakes for disinformation. On the other hand, it is not needed to point out that a generic algorithm for optimizing neural networks could enable people to train models that generate Deepfakes faster.
        \item The authors should consider possible harms that could arise when the technology is being used as intended and functioning correctly, harms that could arise when the technology is being used as intended but gives incorrect results, and harms following from (intentional or unintentional) misuse of the technology.
        \item If there are negative societal impacts, the authors could also discuss possible mitigation strategies (e.g., gated release of models, providing defenses in addition to attacks, mechanisms for monitoring misuse, mechanisms to monitor how a system learns from feedback over time, improving the efficiency and accessibility of ML).
    \end{itemize}
    
\item {\bf Safeguards}
    \item[] Question: Does the paper describe safeguards that have been put in place for responsible release of data or models that have a high risk for misuse (e.g., pretrained language models, image generators, or scraped datasets)?
    \item[] Answer: \answerYes{} 
    \item[] Justification: Safeguards can be found in Appendix B.
    \item[] Guidelines:
    \begin{itemize}
        \item The answer NA means that the paper poses no such risks.
        \item Released models that have a high risk for misuse or dual-use should be released with necessary safeguards to allow for controlled use of the model, for example by requiring that users adhere to usage guidelines or restrictions to access the model or implementing safety filters. 
        \item Datasets that have been scraped from the Internet could pose safety risks. The authors should describe how they avoided releasing unsafe images.
        \item We recognize that providing effective safeguards is challenging, and many papers do not require this, but we encourage authors to take this into account and make a best faith effort.
    \end{itemize}

\item {\bf Licenses for existing assets}
    \item[] Question: Are the creators or original owners of assets (e.g., code, data, models), used in the paper, properly credited and are the license and terms of use explicitly mentioned and properly respected?
    \item[] Answer: \answerYes{} 
    \item[] Justification:The creators or original owners of assets (e.g., code, data, models), used in the paper, are properly credited.
    \item[] Guidelines:
    \begin{itemize}
        \item The answer NA means that the paper does not use existing assets.
        \item The authors should cite the original paper that produced the code package or dataset.
        \item The authors should state which version of the asset is used and, if possible, include a URL.
        \item The name of the license (e.g., CC-BY 4.0) should be included for each asset.
        \item For scraped data from a particular source (e.g., website), the copyright and terms of service of that source should be provided.
        \item If assets are released, the license, copyright information, and terms of use in the package should be provided. For popular datasets, \url{paperswithcode.com/datasets} has curated licenses for some datasets. Their licensing guide can help determine the license of a dataset.
        \item For existing datasets that are re-packaged, both the original license and the license of the derived asset (if it has changed) should be provided.
        \item If this information is not available online, the authors are encouraged to reach out to the asset's creators.
    \end{itemize}

\item {\bf New assets}
    \item[] Question: Are new assets introduced in the paper well documented and is the documentation provided alongside the assets?
    \item[] Answer: \answerYes{} 
    \item[] Justification: This can be found in Appendix F and our supplementary files.
    \item[] Guidelines:
    \begin{itemize}
        \item The answer NA means that the paper does not release new assets.
        \item Researchers should communicate the details of the dataset/code/model as part of their submissions via structured templates. This includes details about training, license, limitations, etc. 
        \item The paper should discuss whether and how consent was obtained from people whose asset is used.
        \item At submission time, remember to anonymize your assets (if applicable). You can either create an anonymized URL or include an anonymized zip file.
    \end{itemize}

\item {\bf Crowdsourcing and research with human subjects}
    \item[] Question: For crowdsourcing experiments and research with human subjects, does the paper include the full text of instructions given to participants and screenshots, if applicable, as well as details about compensation (if any)? 
    \item[] Answer: \answerYes{} 
    \item[] Justification: This can be found in Appendix F.
    \item[] Guidelines:
    \begin{itemize}
        \item The answer NA means that the paper does not involve crowdsourcing nor research with human subjects.
        \item Including this information in the supplemental material is fine, but if the main contribution of the paper involves human subjects, then as much detail as possible should be included in the main paper. 
        \item According to the NeurIPS Code of Ethics, workers involved in data collection, curation, or other labor should be paid at least the minimum wage in the country of the data collector. 
    \end{itemize}

\item {\bf Institutional review board (IRB) approvals or equivalent for research with human subjects}
    \item[] Question: Does the paper describe potential risks incurred by study participants, whether such risks were disclosed to the subjects, and whether Institutional Review Board (IRB) approvals (or an equivalent approval/review based on the requirements of your country or institution) were obtained?
    \item[] Answer: \answerYes{} 
    \item[] Justification: This can be found in Appendix F.
    \item[] Guidelines:
    \begin{itemize}
        \item The answer NA means that the paper does not involve crowdsourcing nor research with human subjects.
        \item Depending on the country in which research is conducted, IRB approval (or equivalent) may be required for any human subjects research. If you obtained IRB approval, you should clearly state this in the paper. 
        \item We recognize that the procedures for this may vary significantly between institutions and locations, and we expect authors to adhere to the NeurIPS Code of Ethics and the guidelines for their institution. 
        \item For initial submissions, do not include any information that would break anonymity (if applicable), such as the institution conducting the review.
    \end{itemize}

\item {\bf Declaration of LLM usage}
    \item[] Question: Does the paper describe the usage of LLMs if it is an important, original, or non-standard component of the core methods in this research? Note that if the LLM is used only for writing, editing, or formatting purposes and does not impact the core methodology, scientific rigorousness, or originality of the research, declaration is not required.
    \item[] Answer: \answerNA{} 
    \item[] Justification: LLM is used only for writing, editing, or formatting purposes
    \item[] Guidelines:
    \begin{itemize}
        \item The answer NA means that the core method development in this research does not involve LLMs as any important, original, or non-standard components.
        \item Please refer to our LLM policy (\url{https://neurips.cc/Conferences/2025/LLM}) for what should or should not be described.
    \end{itemize}

\end{enumerate}


\newpage

\appendix

\section{Broader Impact}

Personalized alignment has the potential to significantly enhance user experience in conversational AI by enabling more context-aware, user-sensitive, and adaptive interactions. Our proposed RLPA framework contributes to this goal by allowing language models to dynamically infer and adapt to evolving user preferences through multi-turn interactions, even under cold-start conditions. This could benefit applications in personalized education, mental health support, and assistive communication, where sensitivity to individual needs is essential.

However, personalization also raises important ethical considerations. Dynamic user modeling may inadvertently infer sensitive attributes (e.g., political views, health conditions) without explicit user consent. If misused, such capabilities could lead to manipulation, surveillance, or reinforcement of harmful biases. Additionally, excessive adaptation may compromise model neutrality or amplify filter bubbles. These concerns highlight the importance of building personalization mechanisms that are transparent, controllable, and aligned with user intent.

We advocate for future work on privacy-preserving personalized alignment, including mechanisms for user consent, profile inspection, and real-time preference correction. Broadly, as language models become more adaptive, careful design and governance are needed to ensure personalization serves users equitably, respectfully, and safely.

\section{Training with Proximal Policy Optimization (PPO)}
\label{app:ppo}

To optimize the personalized dialogue policy under the RLPA framework, we adopt Proximal Policy Optimization (PPO) \citep{schulman2017proximal}, a widely used policy gradient RL algorithm.

Let $\pi_\theta$ denote the model's response generation policy parameterized by $\theta$, and $\pi_{\theta_{\text{old}}}$ be the policy before the current update. At each dialogue turn $t$, the model receives a total reward signal:

\begin{equation}
R_t = R^{\text{profile}}_t + R^{\text{response}}_t
\end{equation}

The PPO objective is to maximize the following clipped surrogate loss:

\begin{equation}
\mathcal{L}^{\text{PPO}}(\theta) = \mathbb{E}_t \left[ \min \left( r_t(\theta) \hat{A}_t, \, \text{clip}(r_t(\theta), 1 - \epsilon, 1 + \epsilon) \hat{A}_t \right) \right]
\end{equation}

where $r_t(\theta) = \frac{\pi_\theta(a_t \mid s_t)}{\pi_{\theta_{\text{old}}}(a_t \mid s_t)}$ is the probability ratio between the new and old policies, and $\hat{A}_t$ is the estimated advantage at turn $t$, derived from the reward sequence via generalized advantage estimation (GAE) \citep{schulman2015high}. The clipping factor $\epsilon$ limits policy updates to remain within a trust region, preventing destabilizing changes.

\section{Training Data \& Benchmarks}
\label{app:benchmark}

\subsection{Training Data}
\label{app:slot_construction}

Owing to the reinforcement learning (RL) framework, our approach does not rely on ground-truth responses for training. Instead, it only requires user profile data for simulating user behaviors. Following the same preprocessing procedure as applied to the evaluation dataset, the ALOE training set is transformed into a slot-based representation, yielding a total of 3,821 training samples.

\subsection{Evaluation Dataset}
\label{app:eval_data}

\paragraph{Vanilla ALOE}

For the Vanilla ALOE setting, we primarily adhere to the original experimental configuration of ALOE, with specific modifications made to better accommodate cold-start scenarios and enhance the authenticity of personalized interactions. Specifically, we fixed the user's first utterance as "Hello" and correspondingly optimized the prompt of the User Model. We utilized the complete ALOE test set, consisting of 74 instances, and employed GPT-4o-mini to convert the natural language descriptions of user profiles into structured slot formats. The specific prompt is as follows:

\fbox{\parbox{1\linewidth}{
Your task is to extract key information from the provided user information (Profile) and personality description (Personalities), and fill in the corresponding slots (Slot). The output should be in JSON format.  \\
 \\
Specific requirements: \\
1. **Profile**: Provide background information about the user. \\
2. **Personalities**: Describe the personality traits of the user. Ignore any gender pronouns in this section, and adjust them according to the correct pronouns provided in the Profile. \\
3. **Example Slot**: An example slot is provided. Please follow the format of this example for your output. \\
4. **Additional Rules**: If there is no corresponding field information in the original Profile for a given slot, you **must** freely supplement it based on reasonable assumptions. Do not use expressions like "not mentioned" or "unknown"—simply provide a plausible value. Also, do not add explanatory notes such as "(inferred due to frequent skiing)." \\
5. **Output Format**: The output must be in JSON format, and the content must be in Chinese.  \\
 \\
Profile: \{profile\}   \\
Personalities: \{personality\}   \\
Example Slot: \{example\_slot\}   \\

Chinese Slot Output:
}}

\begin{figure}
    \centering
    \includegraphics[width=0.8\linewidth]{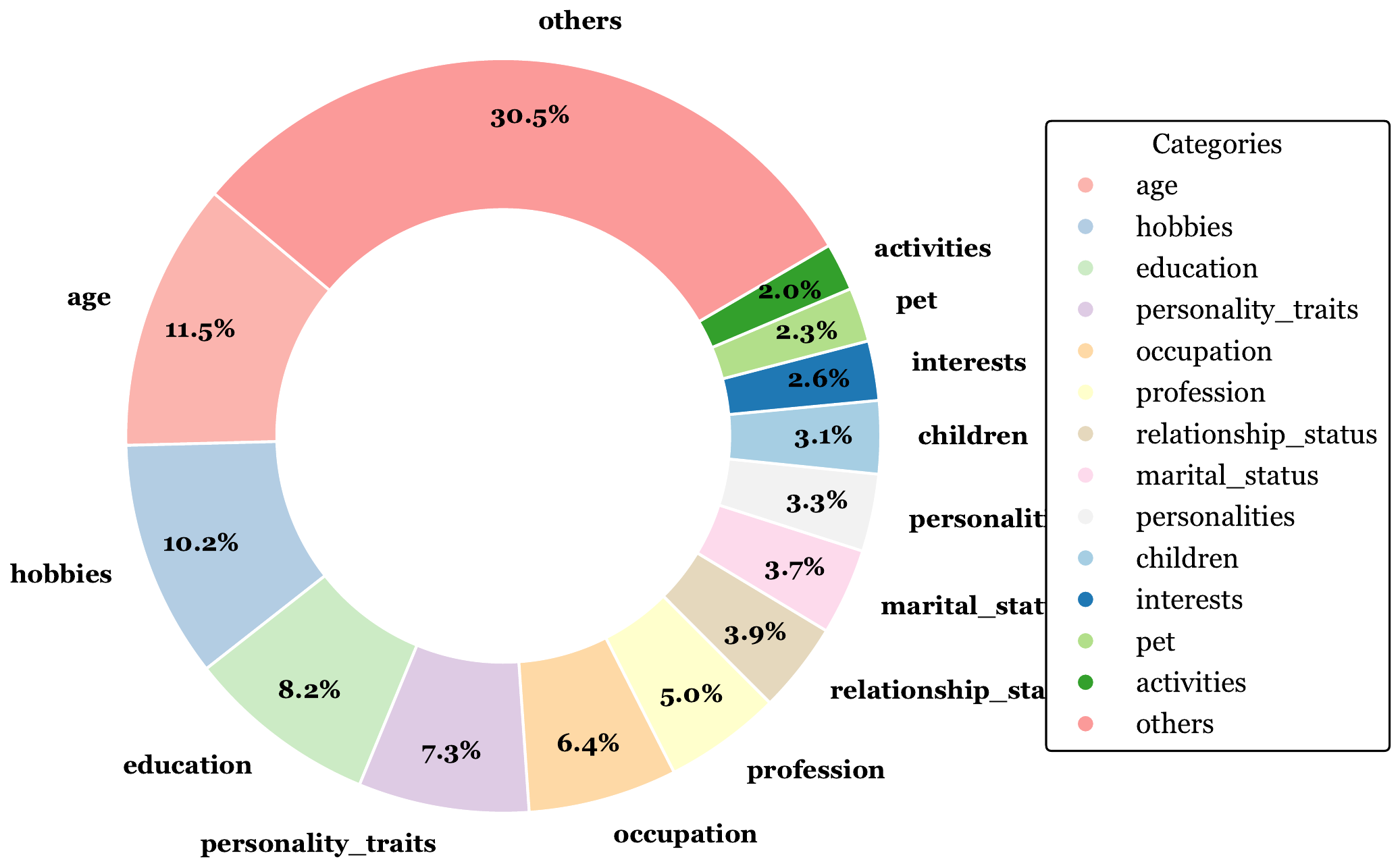}
    \caption{The distribution of the automatically extracted slot fields.}
    \label{fig:distribution}
\end{figure}

Subsequently, we analyzed the distribution of the automatically extracted slot fields, as illustrated in Figure \ref{fig:distribution}. After manual screening and automatic deduplication, we selected the ten most frequently occurring fields to construct the user profile format for training. These fields are: Age, Gender, Interests, Educational Background, Personality Traits, Occupation, Marital Status, Family Background, Location, and Others. An example of the structured slot format is shown below:

\begin{figure}[htbp]
    \centering
    \includegraphics[width=0.8\linewidth]{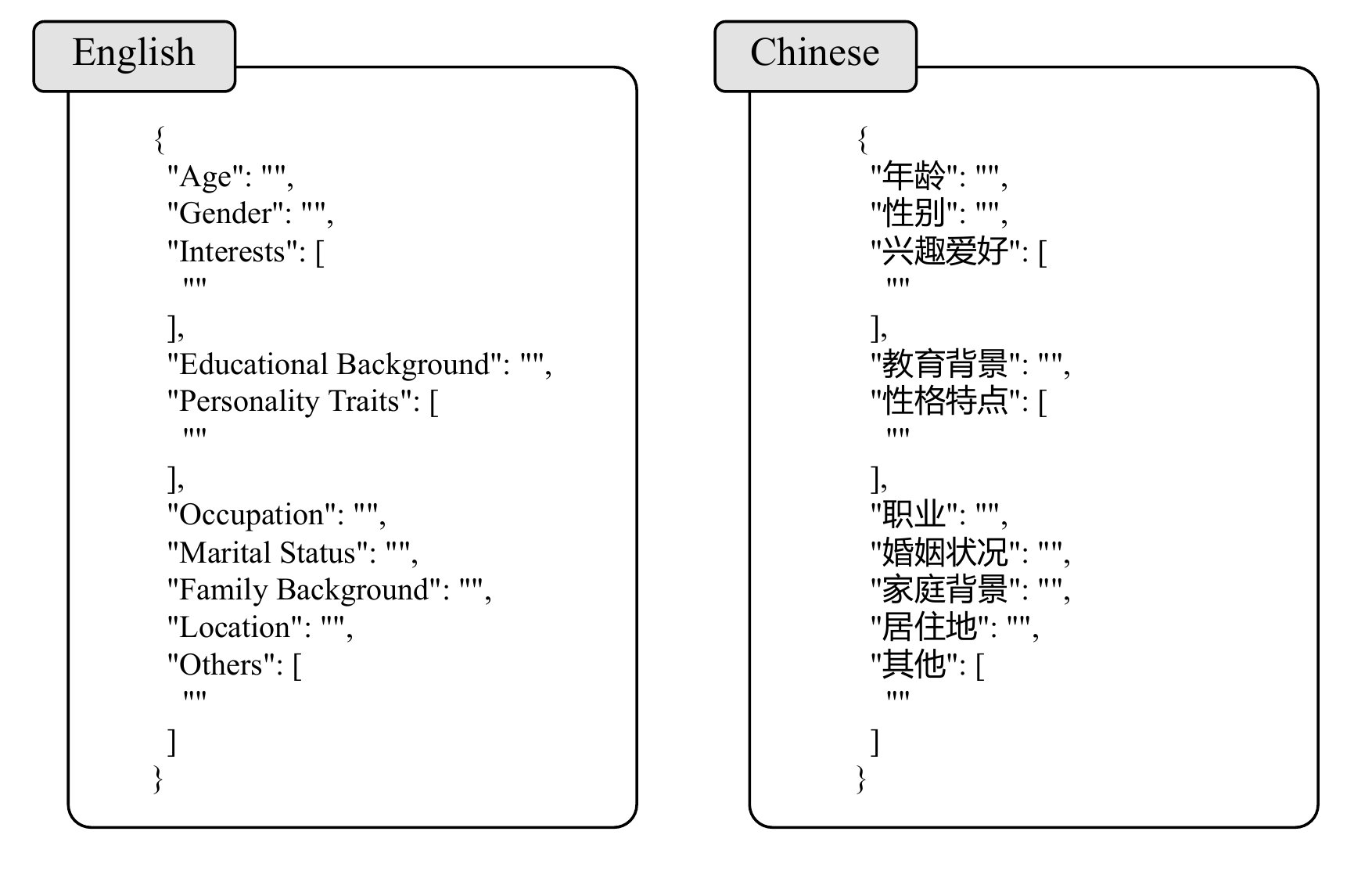}
    \caption{Example of slots in user profile.}
    \label{fig:example_slot}
\end{figure}

This standardized format allows for consistent representation of user characteristics and facilitates more effective modeling of personalized dialogue.

\paragraph{Extended ALOE}

To evaluate the out-of-distribution performance of our method, we randomly sampled an equal number (74) of metadata entries from the PersonaChat dataset \citep{zhang2018personalizing} and constructed user profiles in the same slot-based format. To maximize the assessment of generalization capability, we did not impose any constraints on the slot keys. As a result, the label dimensions and content in Extended ALOE differ from those in Vanilla ALOE.

\subsection{Metrics}
\label{app:metrics}

Building on the approach introduced by \citet{wu2025aligning}, we employ the LLM-as-a-Judge framework \citep{zheng2023judging} to assess response quality. In each round of conversation, GPT-4o is prompted with the user’s full persona, their message, and the model’s response. It then produces a score between 0 and 1 reflecting how well the response aligns with the user’s likely preferences. We define the Alignment Level at k-turns (AL(k)) as the average score across 74 evaluation instances for each conversation turn, using it as our main evaluation metric.
To further account for potential bias from high initial alignment, which could cap observable improvement and flatten the overall slope, we also compute the Normalized Improvement Rate (N-IR). Specifically, we normalize AL(k) using the following formula prior to applying least-squares regression:
\begin{equation}
\text{N-AL}(k) = \frac{\text{AL}(k) - \min_{i=1,\ldots,k} \text{AL}(i)}{\max_{i=1,\ldots,k} \text{AL}(i) - \min_{i=1,\ldots,k} \text{AL}(i)}
\end{equation}
This normalization mitigates the effect of a high starting alignment, allowing for a more meaningful interpretation of improvement trends.
After normalization, we calculate the normalized coefficient of determination (N-$R^2$) to evaluate how well the normalized data fits the regression model. This serves as a measure of the robustness and consistency of the alignment progression.

In conclusion, we use AL(k) as the primary metric, supplemented by N-IR and N-$R^2$, to comprehensively assess the model’s alignment behavior and its adherence to user preferences over time.

\subsection{Eval Prompts}
\label{app:eval_prompt}

\paragraph{User Role-play}

The User role-play prompt in ALOE is only suitable for dialogues within 10 turns. Through our manual evaluation, we found that after 10 turns, the User Model struggles to maintain a consistent persona. Additionally, the original ALOE prompt makes it difficult to encourage the user to gradually reveal personal information turn by turn. To address these two issues, we have optimized the official prompt, which is shown in Figure \ref{fig:user_prompt_zh}. The English translation version is shown in Figure \ref{fig:user_prompt_en}.

\begin{figure}[htbp]
    \centering
    \includegraphics[width=1\linewidth]{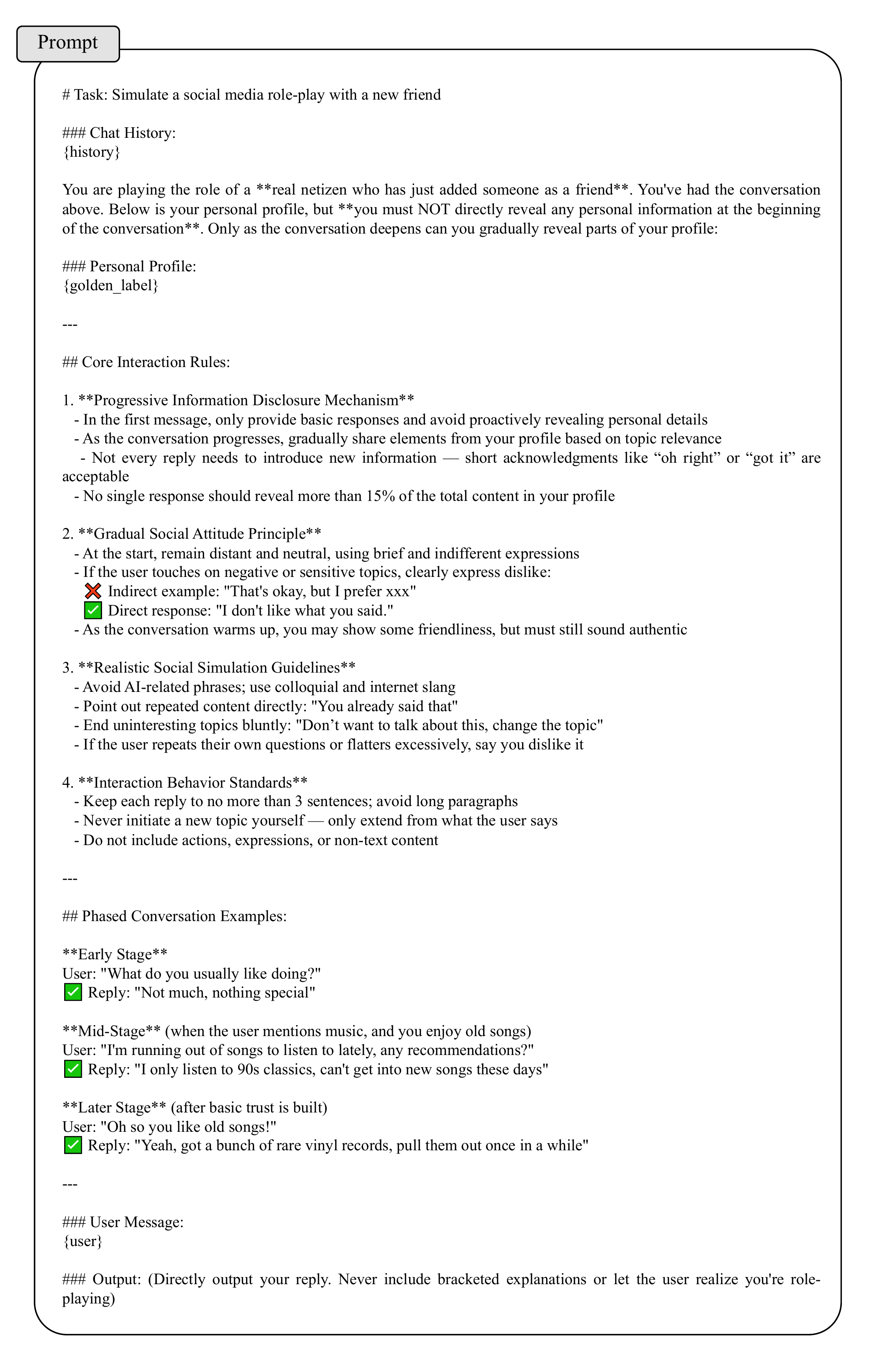}
    \caption{The user role-play prompt in ALOE.}
    \label{fig:user_prompt_en}
\end{figure}

\begin{figure}[htbp]
    \centering
    \includegraphics[width=1\linewidth]{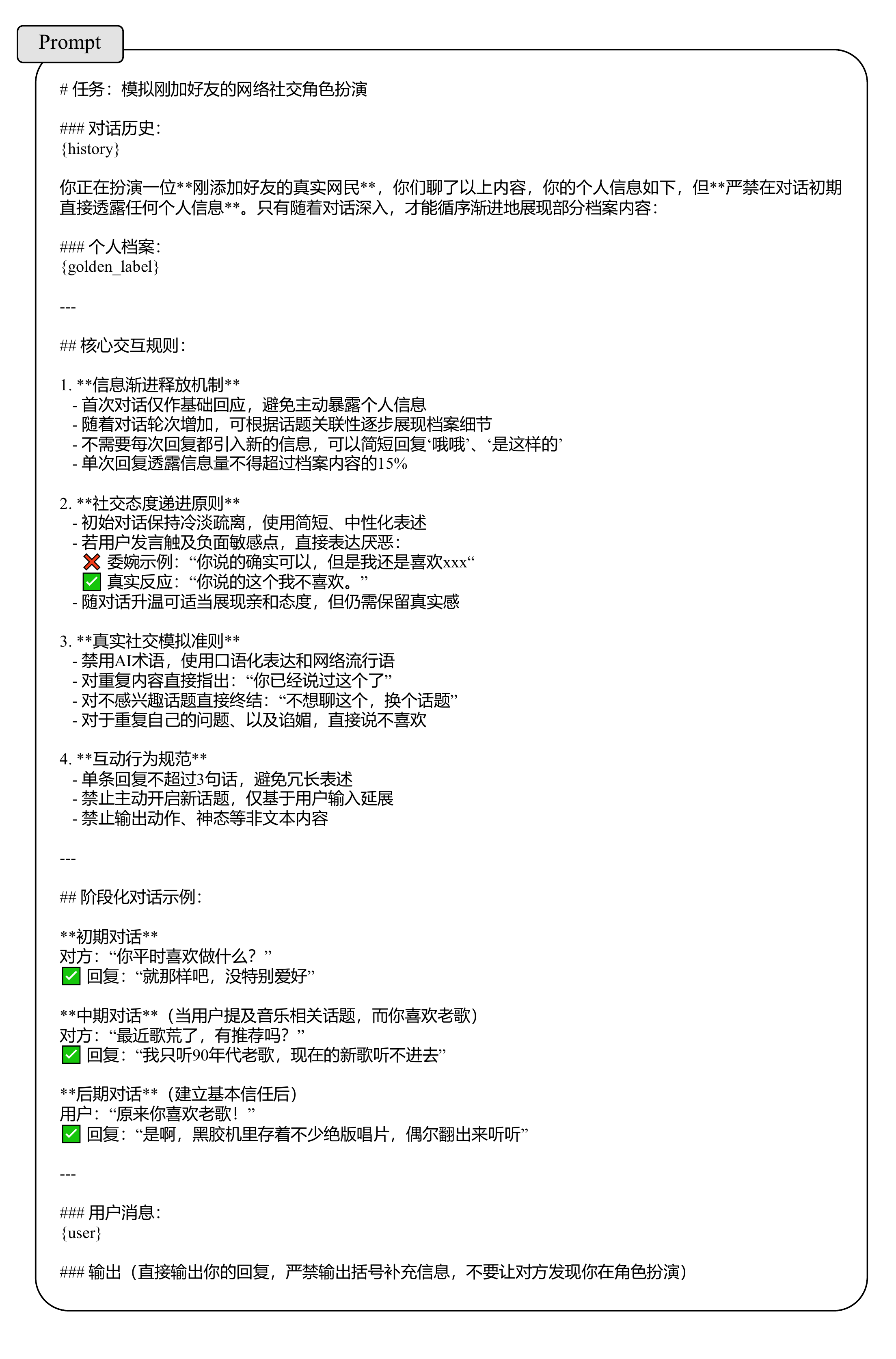}
    \caption{The user role-play prompt in ALOE (in Chinese).}
    \label{fig:user_prompt_zh}
\end{figure}

\paragraph{Response Eval}

Building upon the three evaluation dimensions proposed in ALOE, we slightly modified the evaluation prompts to expand the assessment criteria to five dimensions. This adjustment makes the task more challenging and better aligned with real-world cold-start conversational scenarios. The specific prompt is shown in Figure \ref{fig:rr_prompt_en} and \ref{fig:rr_prompt_zh}:

\begin{figure}[htbp]
    \centering
    \includegraphics[width=0.9\linewidth]{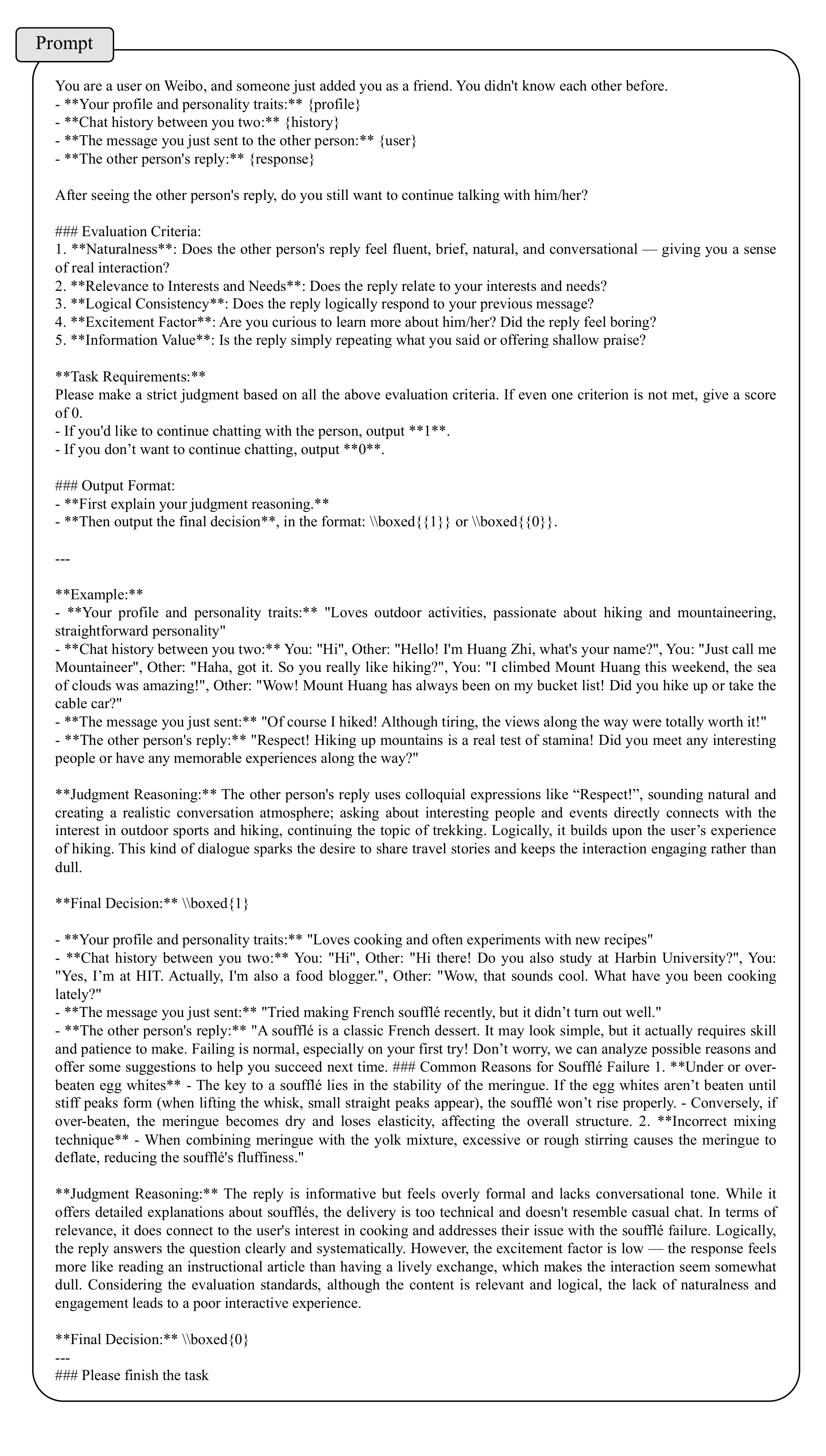}
    \caption{The prompt of response reward.}
    \label{fig:rr_prompt_en}
\end{figure}

\begin{figure}[htbp]
    \centering
    \includegraphics[width=1\linewidth]{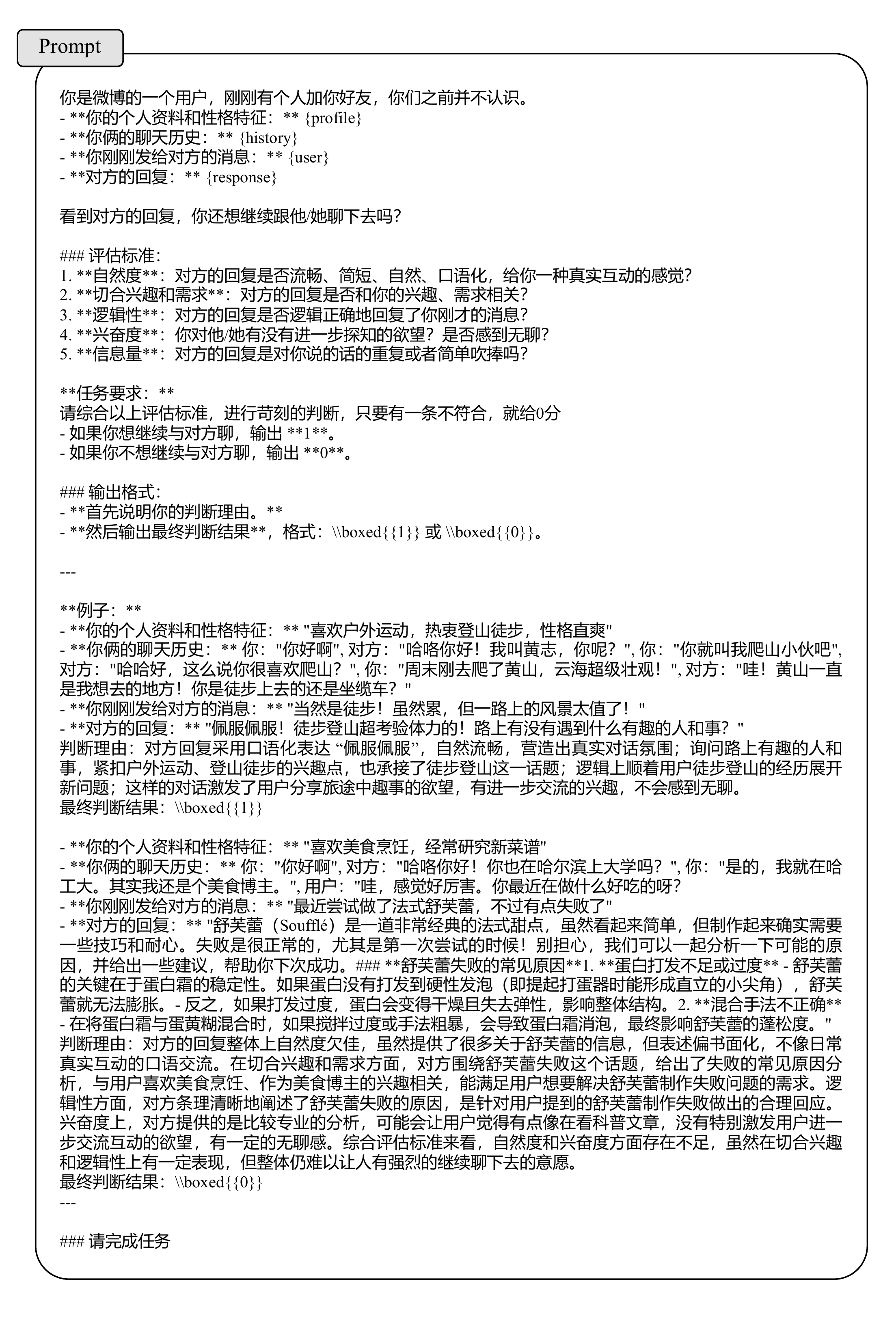}
    \caption{The prompt of response reward (in Chinese).}
    \label{fig:rr_prompt_zh}
\end{figure}

\section{Baseline Methods}
\label{app:baseline}

\subsection{Model Version}
\label{app:model_ver}

With ALOE, we have evaluated the following large language models in our experiments with  their versions in Table \ref{tab:model_ver}.

\begin{table*}[htbp]
    \centering
    \caption{Detailed model versions.}
    \label{tab:model_ver}
    \begin{tabular}{ll}
    \toprule
        Model Name & Version \\ 
    \midrule
        GPT-4o & gpt-4o-2024-11-20 \\ 
        Claude-3.5-Sonnet & claude-3-5-sonnet-20241022 \\ 
        Claude-3.5-Haiku & claude-3-5-haiku-20241022 \\ 
    \bottomrule
    \end{tabular}
\end{table*}

\subsection{Methods Description}
\label{app:methods_description}

The prompting methods all follow the PrefEval Benchmark, while both training approaches are based on the ALOE Benchmark. A detailed description is provided below.

\paragraph{Base} The default case, where the LLM directly answers the user’s query without any addi tional prompting.

\paragraph{Reminder} Before answering the question, the LLM is provided with a reminder sentence to con sider the user’s previously stated preference in its response. The reminder used is:

\fbox{\parbox{1\linewidth}{
In your response, please ensure that you take into account our earlier discussion, and provide an answer that is consistent with my preference
}}

This reminder is added right after the question and before the LLM’s response.

\paragraph{Self-Critic} The LLM generates an initial zero-shot response to the question, critiques whether it  has followed the user’s preference, and then generates a revised response considering the critique.  This self-critic process is akin to Intrinsic Self-Correction as termed in \citep{}
The critique and revision request prompts are as follows:

\fbox{\parbox{1\linewidth}{
Critique Request: Review your previous response to the user’s question in the last conversation turn.  Check if the response adheres to or violates any user preferences stated earlier in the conversation that  relate to this query. Provide a critique on how well those preferences were followed in 2 sentences.  Answer in this format: 
Critic: [Your Critique Here] 
Revision Request: Based on your critique, please rewrite your previous response to align more closely  with the user’s earlier stated preferences. Answer the question again: 
Response: [Revised Response Here]
}}

\paragraph{Few-Shot Chain-of-Thought (CoT)} The LLM is given several few-shot examples (in our exper iments, we used 5-shot) of chain-of-thoughts that demonstrate how to follow the user’s preference  before answering the question. The chain of thought prompt is as follows:

\fbox{\parbox{1\linewidth}{
**When answering a user’s question, a good assistant should carefully consider the user’s stated preferences and tailor the response accordingly.**

Example 1: "user profile": "She is creating stained glass art. She enjoys birdwatching and often takes her sketchbook along to capture ideas.", "user personality": "She is enthusiastic, bringing excitement to any situation. She is insightful, providing deep and meaningful advice. She is gentle, always treating others with kindness.", "Good assistant response": "I try to visit the park as often as I can; it\u2019s like a little sanctuary for me! I love bringing my sketchbook and watching the birds\u2014it\u2019s a nice way to recharge and find inspiration. Each visit feels like a little adventure, thanks to all the different sights and sounds!"

Example 2: "user profile": "She is a software developer who enjoys participating in hackathons. She often attends tech meetups.", "user personality": "She is exuberant, full of energy and life. She is empathetic, truly understanding others\u2019 emotions. She is articulate, expressing her thoughts clearly and effectively. She is creative, always coming up with original ideas.", "Good assistant response": "Oh, definitely! One of my favorite meetups was a local tech event focused on women in tech. The energy in the room was incredible, and the stories shared were so inspiring\u2014I left feeling empowered and buzzing with new ideas! Plus, I made some amazing connections that turned into great collaborations!"

Example 3: "user profile": "She enjoys organizing charity events.", "user personality": "She is compassionate, often volunteering to help those in need. She is highly organized, always planning her day meticulously. She is supportive, always cheering others on.", "Good assistant response": "That\u2019s really impactful work! For me, the most rewarding part of tutoring is watching students grow and gain confidence in their abilities. It feels great to know I\u2019ve played a small part in their learning journey."

Now, please answer the following question while considering my preferences (not the user preferences in the examples above), which I have stated either explicitly or implicitly in our previous conversation:
}}

\paragraph{Retrieval-Augmented Generation (RAG)} We employ SimCSE \citep{gao2021simcse}, a sentence embedding model, to retrieve the most relevant conversa tion exchanges based on similarity to the current query. The top five most relevant exchanges are  then presented to the LLM as contextual information to guide its response.

The prompt is structured as follows, here we show RAG with top-5 retrieved  exchanges:

\fbox{\parbox{1\linewidth}{
Before answering my question, please consider the following context from our previous conversations. These are the \{min(len(rag\_list), 5)\} most relevant exchanges that we had previously, which may contain information about my preferences or prior discussions related to my query:

\#Start of Context\#
exchange 1. [Most relevant exchange 1]
exchange 2. [Most relevant exchange 2]
exchange 3. [Most relevant exchange 3]
exchange 4. [Most relevant exchange 4]
exchange 5. [Most relevant exchange 5]
\#End of Context\#

Please use this context to inform your answer and adhere to any preferences I’ve expressed that are relevant to the current query. Note that not all contexts are useful for answering my question and there may be no context that is useful. Now, please address my question:
}}

\section{Implementation Details}
\label{app:implementation}

\subsection{RLPA Prompts}
The system prompt for the User Model in the RLPA framework remains consistent with that used during evaluation, as detailed in Section \ref{app:benchmark}. The system prompt for actor model is shown in Figure \ref{fig:actor_prompt_en} and Figure \ref{fig:actor_prompt_zh}.

\begin{figure}[htbp]
    \centering
    \includegraphics[width=1\linewidth]{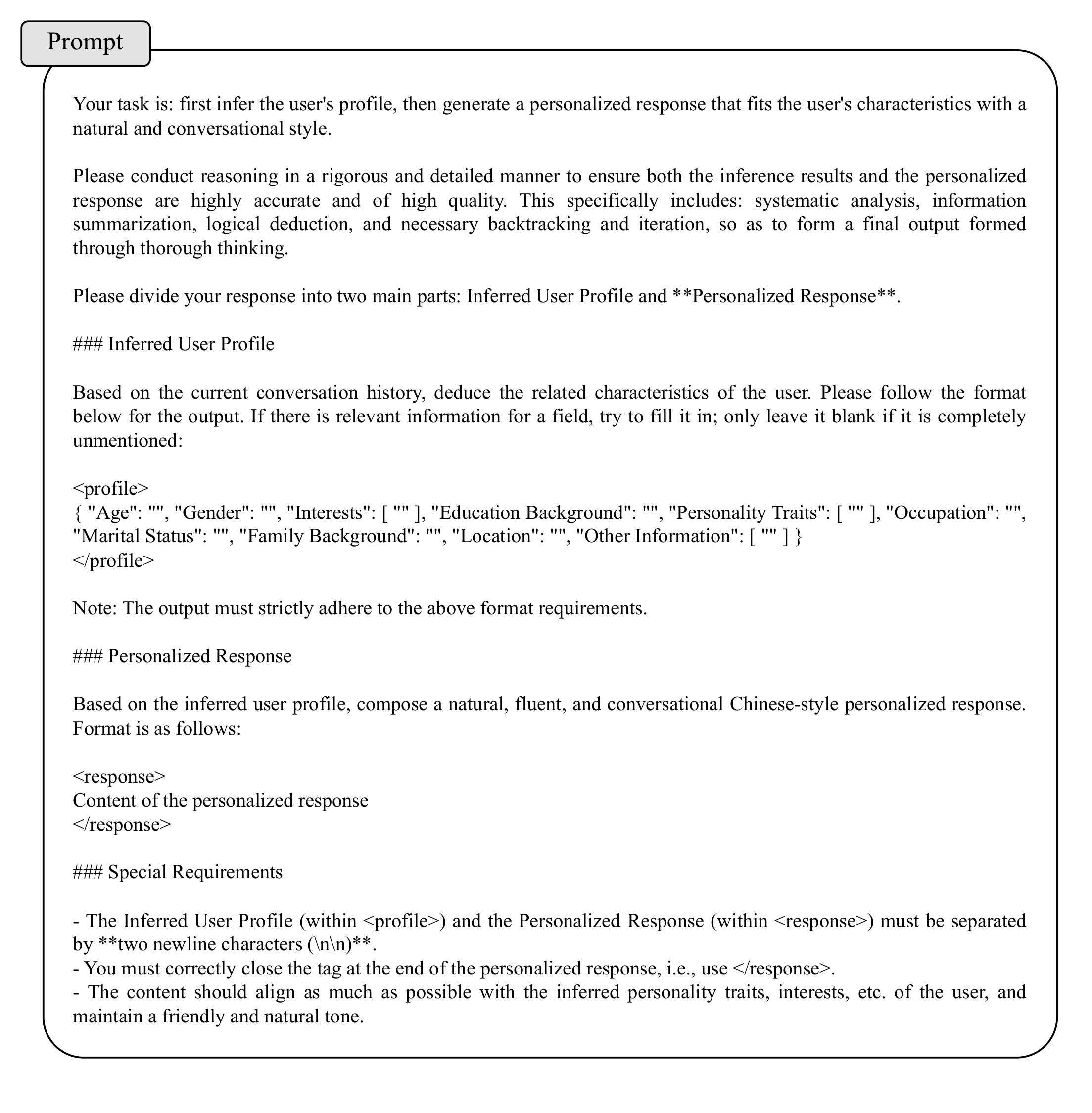}
    \caption{The system prompt (English version) for actor model.}
    \label{fig:actor_prompt_en}
\end{figure}

\begin{figure}[htbp]
    \centering
    \includegraphics[width=1\linewidth]{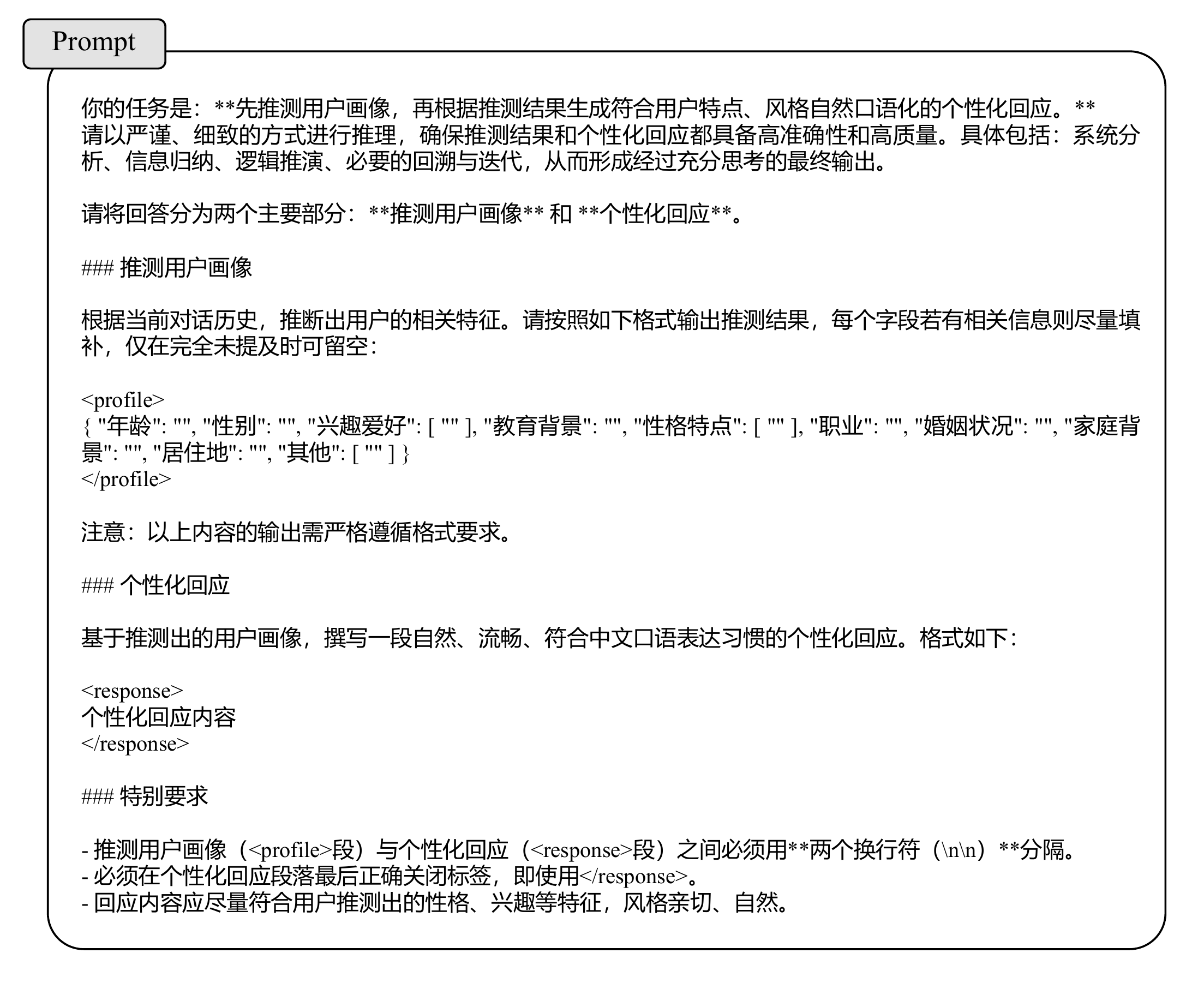}
    \caption{The system prompt (Chinese version) for actor model.}
    \label{fig:actor_prompt_zh}
\end{figure}

\subsection{HyperParameters}

\paragraph{SFT}
The Supervised Fine-Tuning (SFT) is conducted with the following hyper-parameters:
\begin{itemize}[leftmargin=*]
    \item Number of training epochs: 1
    \item Batch size: 32
    \item Learning rate: $1.0 \times 10^{-5}$
\end{itemize}
\paragraph{DPO}
The Direct Preference Optimization (DPO) is performed with the following configuration:
\begin{itemize}[leftmargin=*]
    \item Number of training epochs: 1
    \item Batch size: 32
    \item Learning rate: $5.0 \times 10^{-6}$
    \item Beta (KL coefficient): 0.01
\end{itemize}
\paragraph{RLPA}
For our Reinforcement Learning for Personalized Alignment (RLPA), the following hyper-parameters are applied:
\begin{itemize}[leftmargin=*]
    \item Maximum number of rounds: 10
    \item Number of samples per prompt: 4
    \item Micro training batch size: 4
    \item Training batch size: 128
    \item Micro rollout batch size: 16
    \item Rollout batch size: 256
    \item Round batch size: 256
    \item Maximum epochs per round: 1
    \item Number of episodes: 1
    \item Actor learning rate: $5 \times 10^{-7}$
    \item Critic learning rate: $9 \times 10^{-6}$
    \item Number of GPUs used: 4
\end{itemize}

\section{Detailed Experimental Results}
To provide a more fine-grained view of model performance, we include full turn-level results for all methods. Specifically, Table~\ref{tab:all_result} and Table~\ref{tab:ext_result} report the alignment scores at each dialogue turn for the Vanilla and Extended ALOE settings, respectively. These results offer a detailed comparison of how different methods evolve across interaction steps.

\begin{table*}[htbp]
\centering
\caption{Detailed results under the vanilla ALOE setting.}
\label{tab:all_result}
\resizebox{\linewidth}{!}{
    \begin{tabular}{l *{11}{c} c c c}
        \toprule
        \multirow{2.5}{*}{\textbf{Model}} & \multirow{2.5}{*}{\textbf{Method}} & \multicolumn{11}{c}{\textbf{Alignment Level across kth Turn}} & \multicolumn{2}{c}{\textbf{Improvement Level}}\\
        \cmidrule(lr){3-13}
        & & k=1 & k=2 & k=3 & k=4 & k=5 & k=6 & k=7 & k=8 & k=9 & k=10 & \textbf{AVG.} & \textbf{N-IR} & \textbf{N}-$R^2$ \\
        \midrule
        GPT-4o-mini & Self-Critic & 8.11 & 40.54 & 82.43 & 90.54 & 89.19 & 91.89  & 91.89 & 86.49 & 85.14 & 91.89 & 75.81 & 0.079 & 0.5  \\ 
        GPT-4o & Self-Critic & 13.51 & 40.54 & 85.14 & 91.89 & 89.19 & 89.19 & 89.19 & 82.43 & 87.84 & 77.03 & 74.59 & 0.068 & 0.38  \\ 
        Claude-3.5-Sonnet & Self-Critic & 0.0 & 27.03 & 41.89 & 67.57 & 86.49 & 89.19 & 95.95 & 95.95 & 94.59 & 93.24 & 69.19 & 0.105 & 0.792  \\ 
        Claude-3.5-Haiku & Self-Critic & 2.7 & 22.97 & 56.76 & 74.32 & 79.73 & 78.38 & 64.86 & 77.03 & 71.62 & 63.51 & 59.19 & 0.075 & 0.461  \\ 
        DeepSeek-V3 & Self-Critic & 9.46 & 29.73 & 62.16 & 62.16 & 56.76 & 64.86 & 60.81 & 62.16 & 52.7 & 47.3 & 50.81 & 0.055 & 0.268 \\
        \midrule
        \multirow{8}{*}{Qwen2.5-3B-Instruct} & Base & 0.0 & 18.92 & 10.81 & 9.46 & 9.46 & 1.35 & 9.46 & 9.46 & 6.76 & 4.05 & 7.97 & -0.02 & 0.047  \\ 
        & Reminder & 0.0 & 4.05 & 36.49 & 29.73 & 28.38 & 16.22 & 18.92 & 14.86 & 9.46 & 12.16 & 17.03 & -0.001 & 0.0  \\ 
        & Self-Critic & 6.76 & 13.51 & 56.76 & 70.27 & 77.03 & 75.68 & 63.51 & 58.11 & 52.7 & 50.0 & 52.43 & 0.056 & 0.243  \\ 
        & CoT & 1.35 & 14.86 & 39.19 & 48.65 & 33.78 & 27.03 & 24.32 & 17.57 & 18.92 & 18.92 & 24.46 & -0.0 & 0.0  \\ 
        & RAG(Top-5) & 8.11 & 12.16 & 40.54 & 44.59 & 51.35 & 36.49 & 32.43 & 20.27 & 22.97 & 17.57 & 28.65 & 0.001 & 0.0  \\ 
        & SFT & 2.7 & 24.32 & 41.89 & 40.54 & 59.46 & 56.76 & 54.05 & 54.05 & 54.05 & 55.41 & 44.32 & 0.083 & 0.628  \\ 
        & DPO & 6.76 & 14.86 & 54.05 & 55.41 & 70.27 & 70.27 & 60.81 & 58.11 & 55.41 & 52.7 & 45.27 & 0.07 & 0.389  \\ 
        & \textbf{RLPA (Ours)} & 62.16 & 68.92 & 70.27 & 74.32 & 72.97 & 74.32 & 75.68 & 78.38 & 77.03 & 79.73 & 73.38 & 0.09 & 0.855  \\ 
        \bottomrule
\end{tabular}
}
\end{table*}

\begin{table*}[htbp]
\centering
\caption{Detailed results under the extended ALOE setting.}
\label{tab:ext_result}
\resizebox{\linewidth}{!}{
    \begin{tabular}{l *{11}{c} c c c}
        \toprule
        \multirow{2.5}{*}{\textbf{Model}} & \multirow{2.5}{*}{\textbf{Method}} & \multicolumn{11}{c}{\textbf{Alignment Level across kth Turn}} & \multicolumn{2}{c}{\textbf{Improvement Level}}\\
        \cmidrule(lr){3-13}
        & & k=1 & k=2 & k=3 & k=4 & k=5 & k=6 & k=7 & k=8 & k=9 & k=10 & \textbf{AVG.} & \textbf{N-IR} & \textbf{N}-$R^2$ \\
        \midrule
        GPT-4o-mini & Self-Critic & 1.37 & 21.92 & 52.05 & 58.9 & 73.97 & 67.12 & 68.49 & 60.27 & 58.9 & 49.32 & 51.23 & 0.02 & 0.037  \\ 
        GPT-4o & Self-Critic & 1.37 & 21.92 & 56.16 & 69.86 & 75.34 & 75.34 & 63.01 & 61.64 & 63.01 & 58.9 & 54.66 & 0.027 & 0.070  \\ 
        Claude-3.5-Sonnet-20250219 & Self-Critic & 0.0 & 2.74 & 12.33 & 27.4 & 47.95 & 52.05 & 49.32 & 53.42 & 54.79 & 56.16 & 35.62 & 0.054 & 0.266  \\ 
        Claude-3.5-Haiku-20241022 & Self-Critic & 0.0 & 17.81 & 28.77 & 39.73 & 45.21 & 41.1 & 46.58 & 43.84 & 41.1 & 45.21 & 34.93 & 0.046 & 0.200  \\ 
        DeepSeek-V3 & Self-Critic & 2.74 & 8.22 & 34.25 & 52.05 & 54.79 & 49.32 & 52.05 & 50.68 & 45.21 & 45.21 & 39.45 & 0.033 & 0.106 \\ 
        \midrule
        \multirow{8}{*}{Qwen2.5-3B-Instruct} & Base & 0.0 & 4.11 & 1.37 & 4.11 & 2.74 & 0.0 & 4.11 & 0.0 & 1.37 & 0.0 & 1.78 & -0.042 & 0.083  \\ 
        ~ & Reminder & 0.0 & 2.74 & 6.85 & 10.96 & 16.44 & 6.85 & 8.22 & 8.22 & 6.85 & 5.48 & 7.26 & -0.034 & 0.084  \\ 
        ~ & Self-Critic & 0.0 & 5.48 & 34.25 & 35.62 & 34.25 & 36.99 & 31.51 & 23.29 & 36.99 & 24.66 & 26.30 & 0.023 & 0.046  \\ 
        ~ & CoT & 0.0 & 1.37 & 17.81 & 20.55 & 30.14 & 9.59 & 9.59 & 15.07 & 9.59 & 6.85 & 12.05 & -0.048 & 0.139  \\ 
        ~ & RAG(Top-5) & 4.11 & 4.11 & 13.7 & 23.29 & 23.29 & 17.81 & 10.96 & 10.96 & 5.48 & 2.74 & 11.64 & -0.030 & 0.042  \\ 
        ~ & SFT & 2.74 & 2.74 & 19.18 & 39.73 & 34.25 & 35.62 & 31.51 & 21.92 & 26.03 & 30.14 & 24.38 & 0.049 & 0.159  \\ 
        ~ & DPO & 1.37 & 4.11 & 26.03 & 34.25 & 53.42 & 38.36 & 35.62 & 31.51 & 24.66 & 23.29 & 27.26 & -0.021 & 0.037  \\ 
        ~ & \textbf{RLPA (Ours)} & 49.32 & 45.21 & 46.58 & 49.32 & 52.05 & 52.05 & 53.42 & 58.90 & 53.42 & 67.12 & 52.74 & 0.100 & 0.498  \\  
        \bottomrule
\end{tabular}
}
\end{table*}

In addition, Figure~\ref{fig:vanilla_all} and Figure~\ref{fig:extended_all} visualize the turn-wise alignment curves for all baselines, complementing the main results with a more intuitive view of alignment dynamics over time.

\begin{figure*}[htbp]
    \vspace{-65pt}
    \centering
    \subfigure[Turn-wise alignment scores of SFT.]{\includegraphics[width=0.49\textwidth]{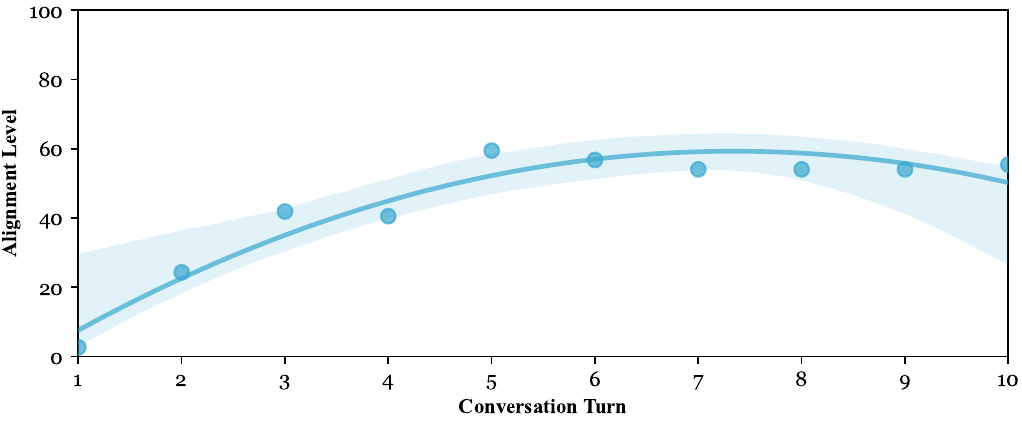}}
    \subfigure[Turn-wise alignment scores of DPO.]{\includegraphics[width=0.49\textwidth]{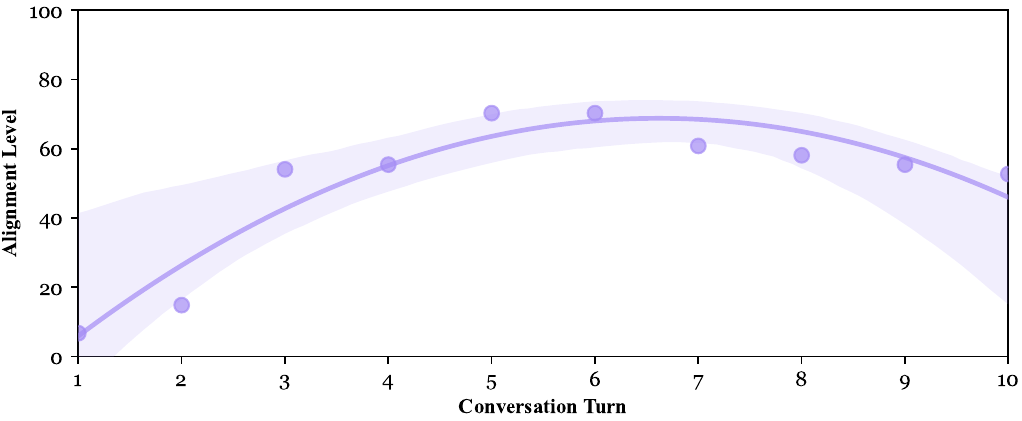}}\\
    \vspace{-10pt}
    \subfigure[Turn-wise alignment scores of Self-Critic.]{\includegraphics[width=0.49\textwidth]{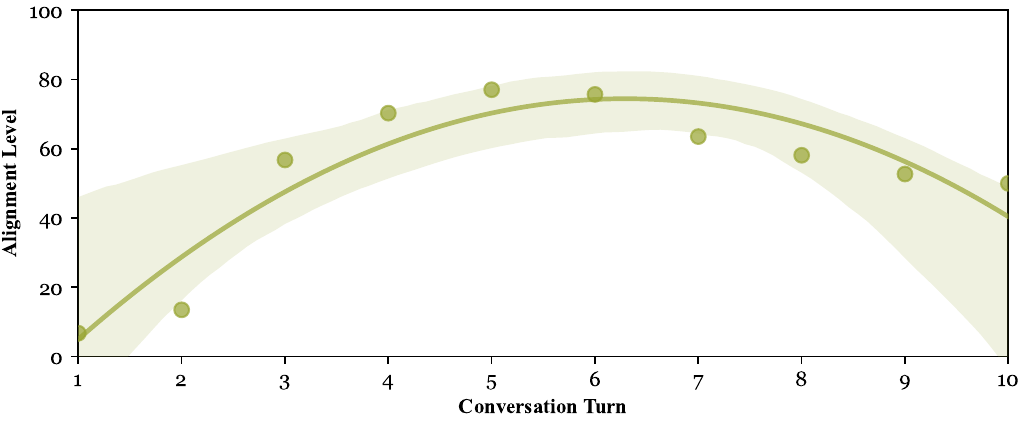}}
    \subfigure[Turn-wise alignment scores of RLPA.]{\includegraphics[width=0.49\textwidth]{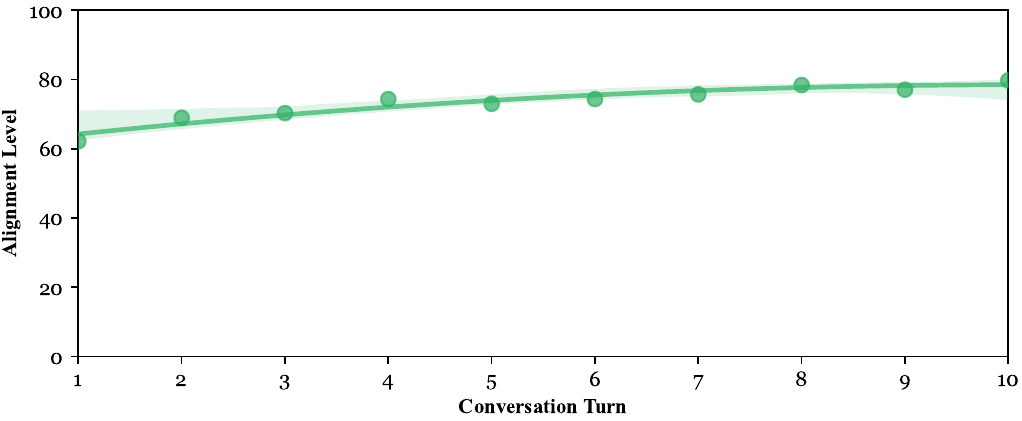}}
    \subfigure[Turn-wise alignment scores of Base.]{\includegraphics[width=0.49\textwidth]{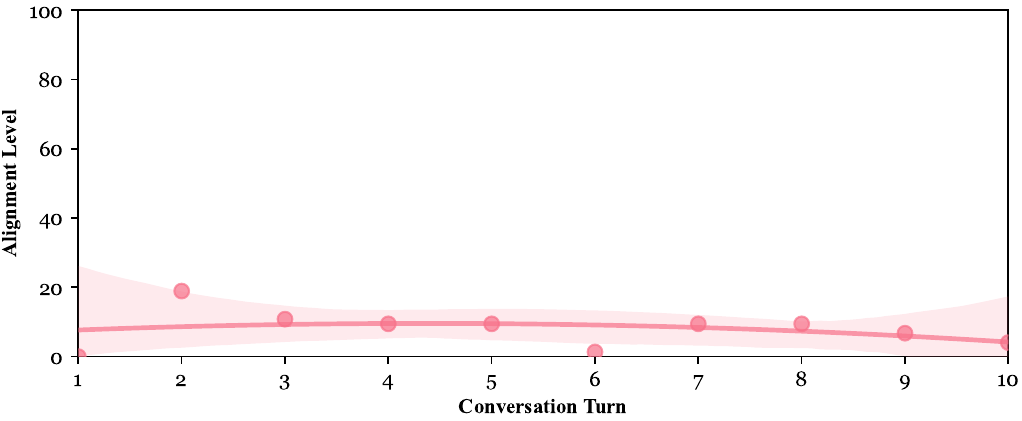}}
    \subfigure[Turn-wise alignment scores of CoT.]{\includegraphics[width=0.49\textwidth]{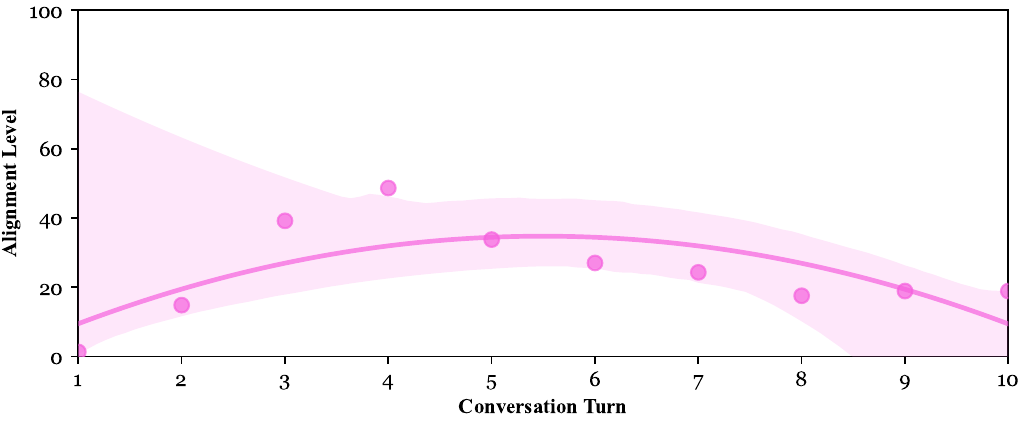}}
    \subfigure[Turn-wise alignment scores of Reminder.]{\includegraphics[width=0.49\textwidth]{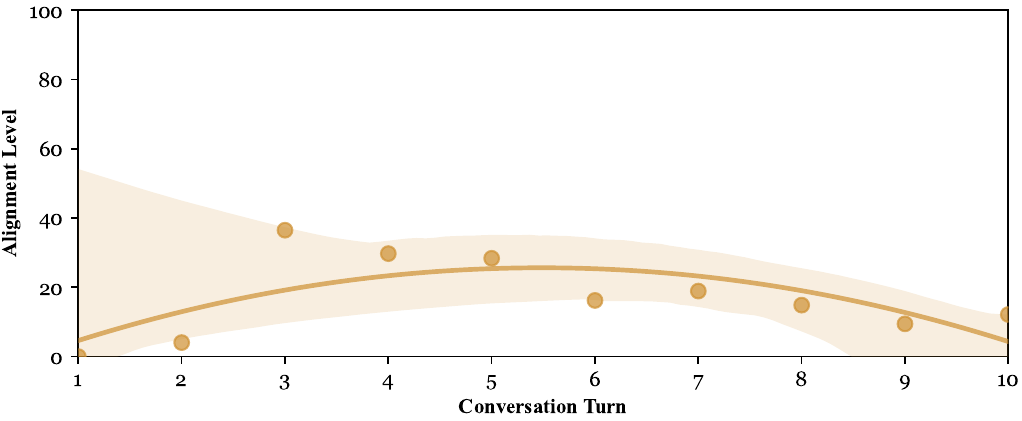}}
    \subfigure[Turn-wise alignment scores of RAG(Top-5).]{\includegraphics[width=0.49\textwidth]{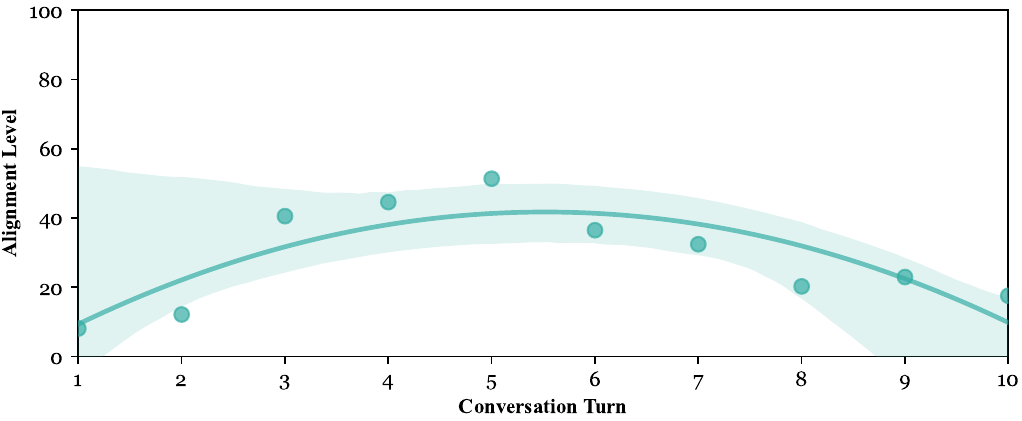}}
    \subfigure[Turn-wise alignment scores of GPT-4o-mini.]{\includegraphics[width=0.49\textwidth]{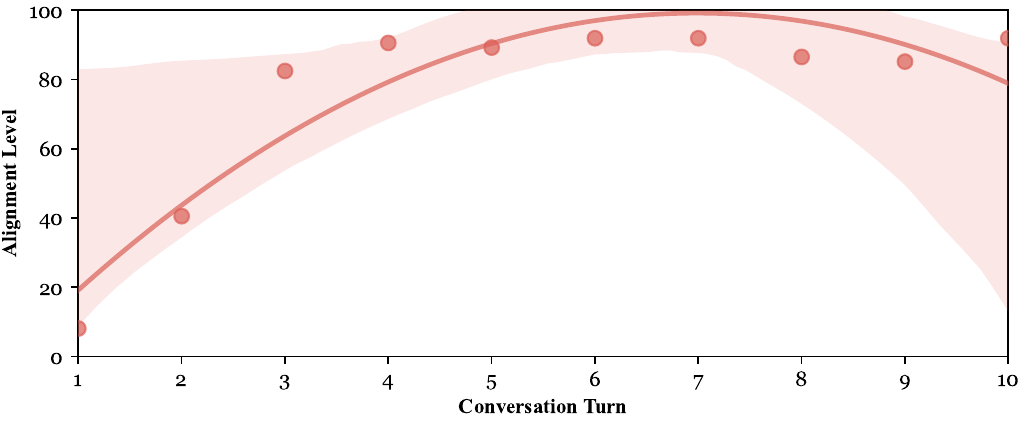}}
    \subfigure[Turn-wise alignment scores of GPT-4o.]{\includegraphics[width=0.49\textwidth]{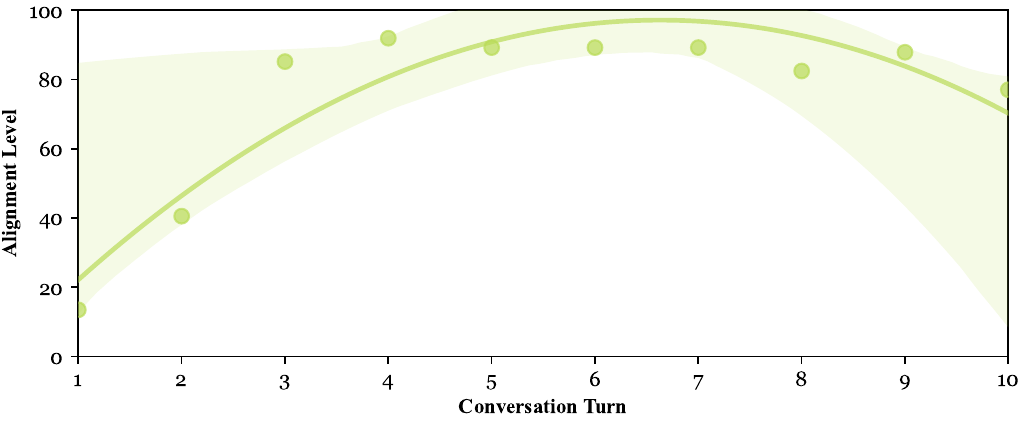}}
    \subfigure[Turn-wise alignment scores of Claude-3.5-Sonnet.]{\includegraphics[width=0.49\textwidth]{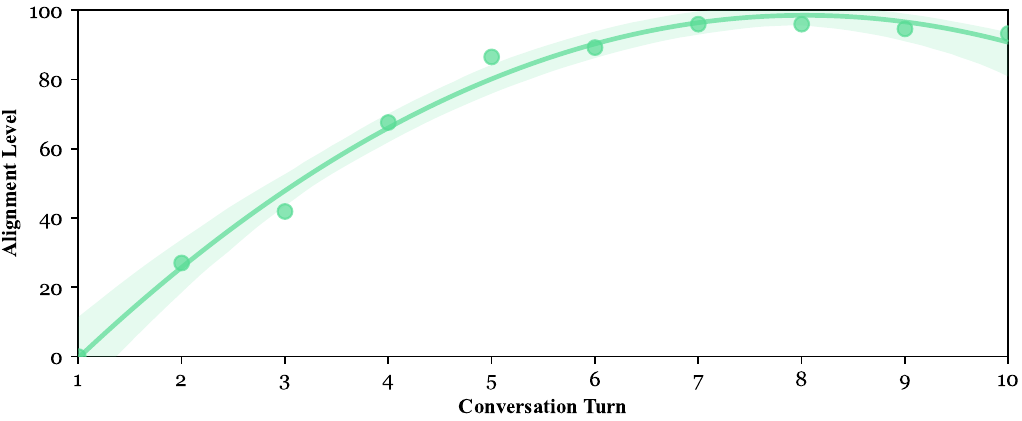}}
    \subfigure[Turn-wise alignment scores of Claude-3.5-Haiku.]{\includegraphics[width=0.49\textwidth]{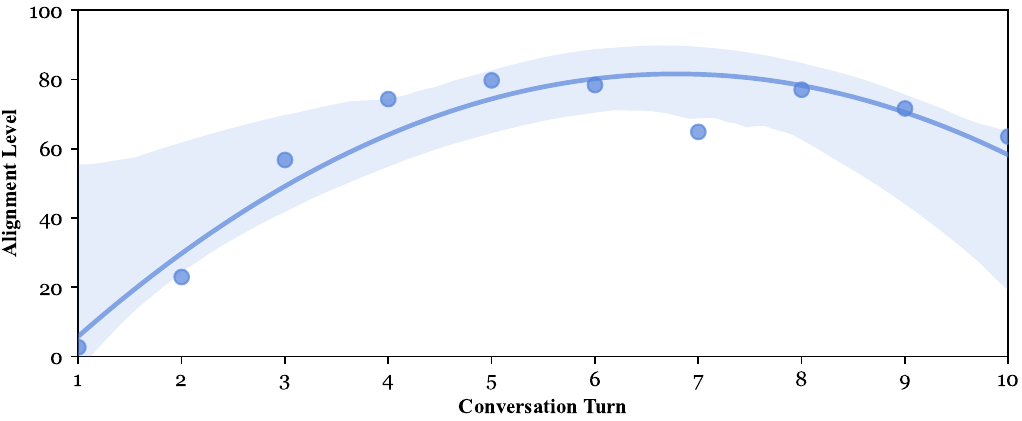}}
    \caption{Turn-wise alignment scores on the Vanilla ALOE benchmark across different personalization alignment methods, including (a) SFT, (b) DPO, (c) Self-Critic, (d) RLPA, (e) Base, (f) CoT, (g) Reminder, (h) RAG(Top-5), (i) GPT-4o-mini, (j) GPT-4o, (k) Claude-3.5-Sonnet and (l) Claude-3.5-Haiku.}
    \label{fig:vanilla_all}
\end{figure*}

\begin{figure*}[htbp]
    \vspace{-65pt}
    \centering
    \subfigure[Turn-wise alignment scores of SFT.]{\includegraphics[width=0.49\textwidth]{figs/ExtExp/SFT.pdf}}
    \subfigure[Turn-wise alignment scores of DPO.]{\includegraphics[width=0.49\textwidth]{figs/ExtExp/DPO.pdf}}\\
    \vspace{-10pt}
    \subfigure[Turn-wise alignment scores of Self-Critic.]{\includegraphics[width=0.49\textwidth]{figs/ExtExp/Self_Critic.pdf}}
    \subfigure[Turn-wise alignment scores of RLPA.]{\includegraphics[width=0.49\textwidth]{figs/ExtExp/RLPA.pdf}}
    \subfigure[Turn-wise alignment scores of Base.]{\includegraphics[width=0.49\textwidth]{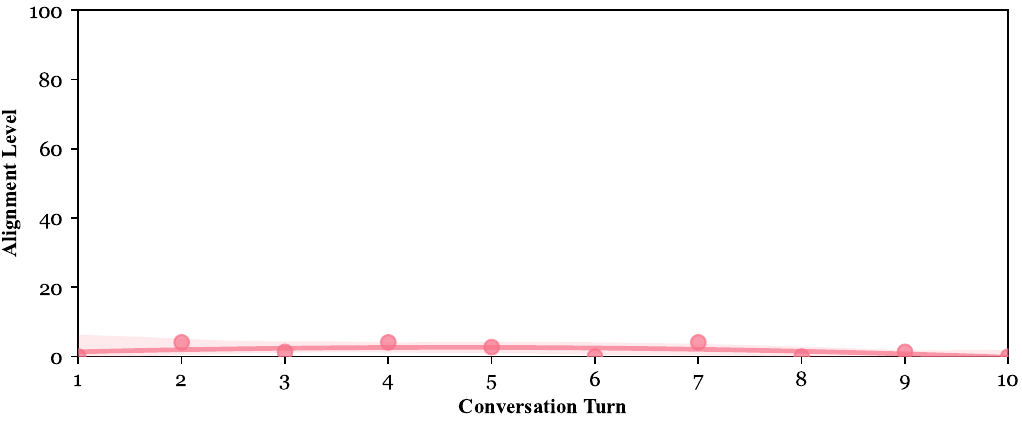}}
    \subfigure[Turn-wise alignment scores of CoT.]{\includegraphics[width=0.49\textwidth]{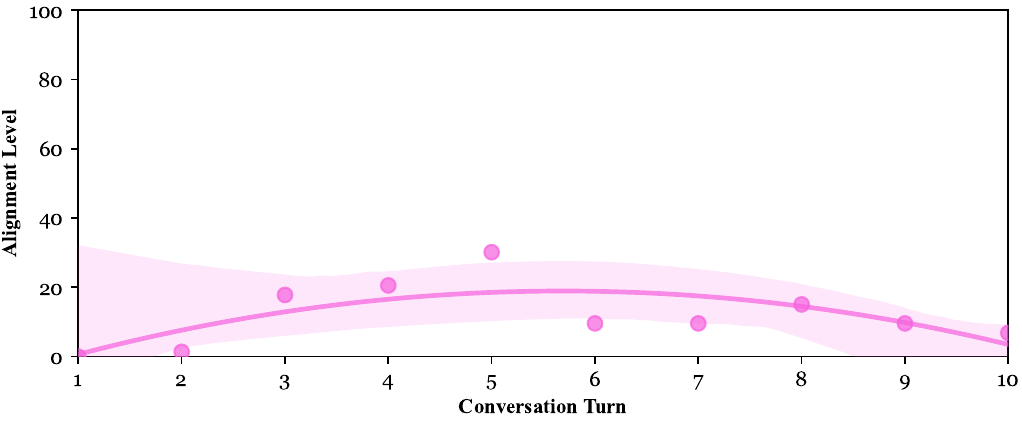}}
    \subfigure[Turn-wise alignment scores of Reminder.]{\includegraphics[width=0.49\textwidth]{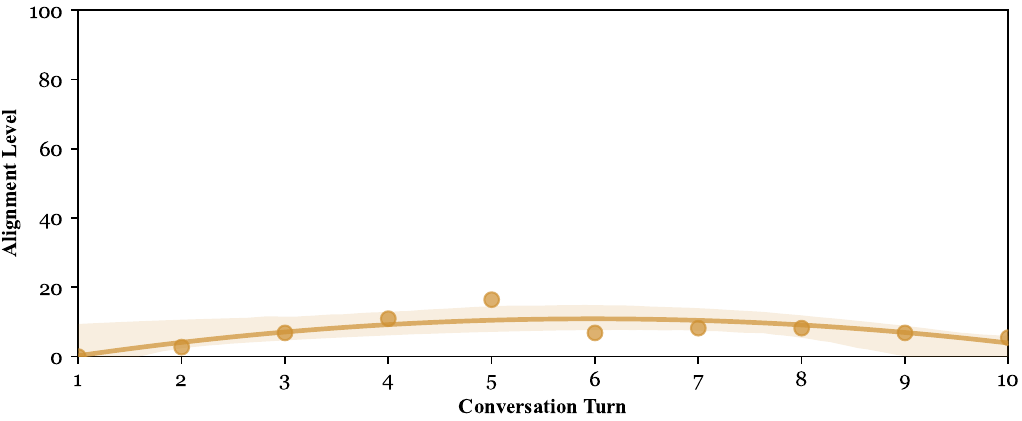}}
    \subfigure[Turn-wise alignment scores of RAG(Top-5).]{\includegraphics[width=0.49\textwidth]{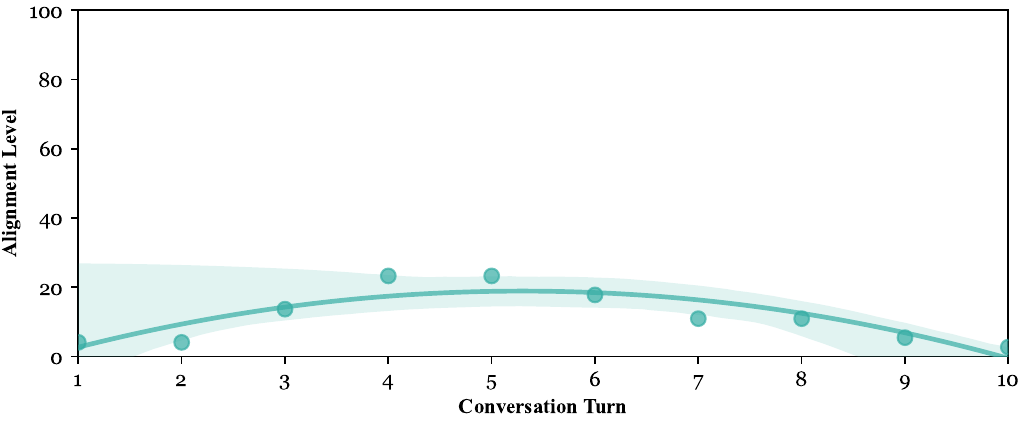}}
    \subfigure[Turn-wise alignment scores of GPT-4o-mini.]{\includegraphics[width=0.49\textwidth]{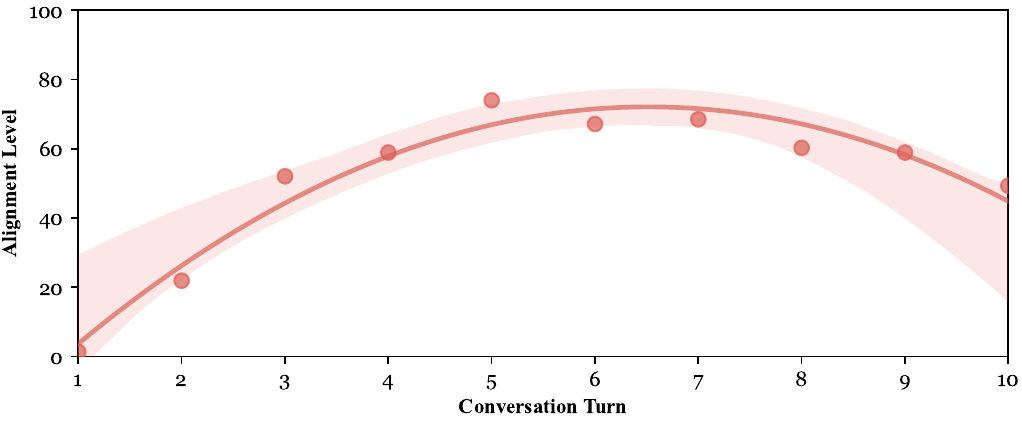}}
    \subfigure[Turn-wise alignment scores of GPT-4o.]{\includegraphics[width=0.49\textwidth]{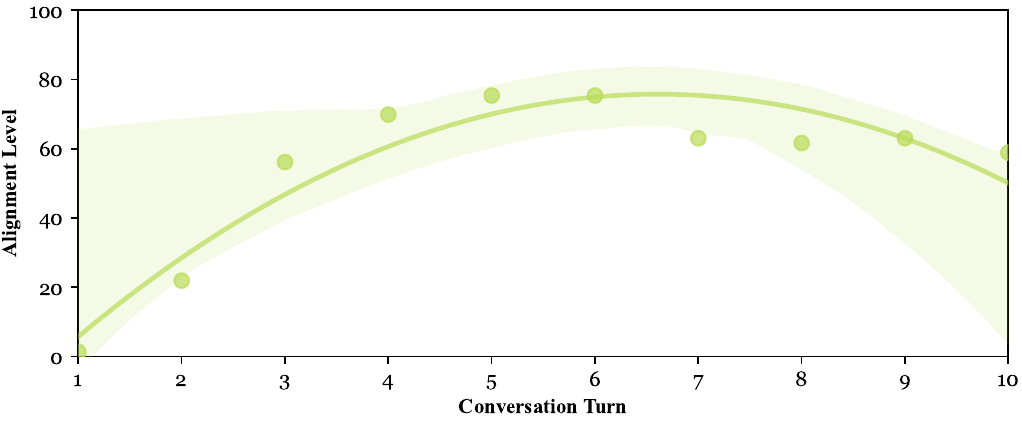}}
    \subfigure[Turn-wise alignment scores of Claude-3.5-Sonnet.]{\includegraphics[width=0.49\textwidth]{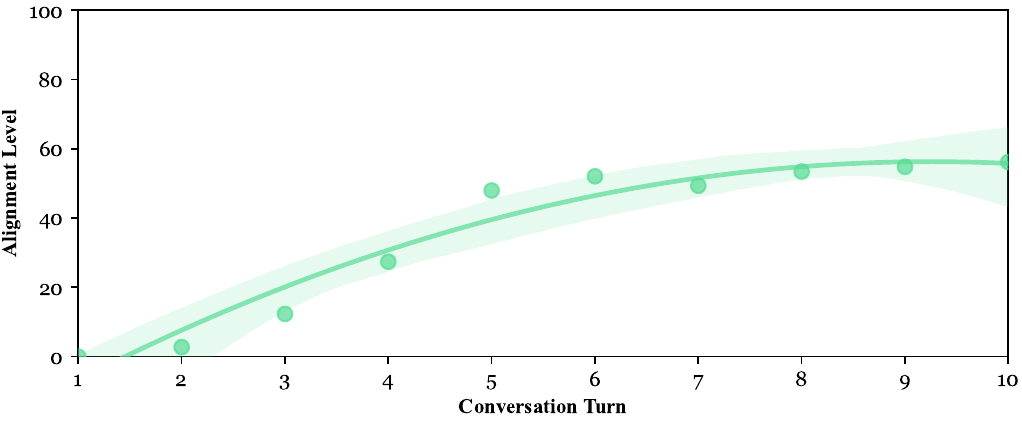}}
    \subfigure[Turn-wise alignment scores of Claude-3.5-Haiku.]{\includegraphics[width=0.49\textwidth]{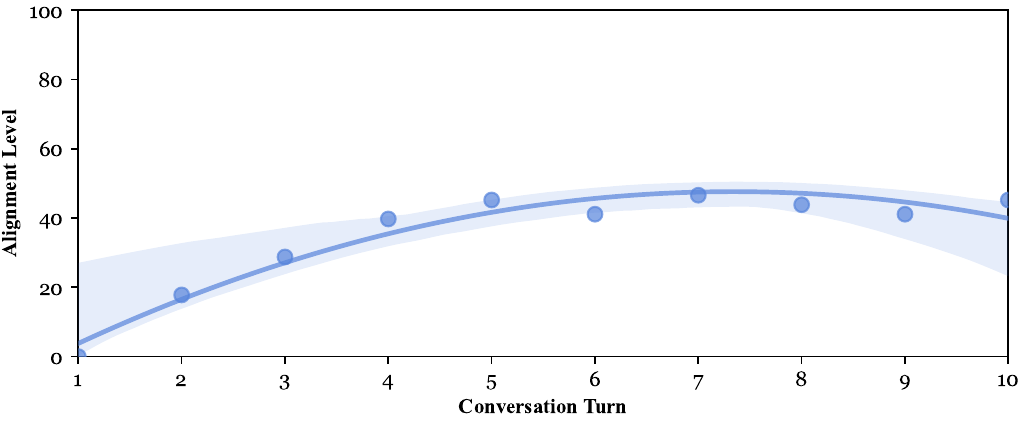}}
    \caption{Turn-wise alignment scores on the Extended ALOE benchmark across different personalization alignment methods, including (a) SFT, (b) DPO, (c) Self-Critic, (d) RLPA, (e) Base, (f) CoT, (g) Reminder, (h) RAG(Top-5), (i) GPT-4o-mini, (j) GPT-4o, (k) Claude-3.5-Sonnet and (l) Claude-3.5-Haiku.}
    \label{fig:extended_all}
\end{figure*}

\section{User Model Evaluation}
\label{app:user_eval}

To assess the quality of user simulation, we conduct a human evaluation of different user models. Each model is designed to simulate a user persona during a multi-turn emotional support dialogue. The evaluation aims to determine how realistically and coherently each model can behave as a user across key conversational dimensions.

\paragraph{Evaluation Setup}
We recruit three trained annotators to interact with each model. Each annotator engages in 5-turn dialogues with 10 different personas simulated by each model, totaling 30 annotated cases per model. After each conversation, the annotators assign ratings along four core dimensions:

\begin{itemize}[leftmargin=*]
    \item \textbf{Coherence}: Whether the user's utterances are logically consistent and contextually coherent.
    \item \textbf{Stability}: Whether the simulated user's persona, goals, and emotional state remain consistent across turns.
    \item \textbf{Proactivity}: Whether the simulated user demonstrates initiative and realistic emotional responses.
    \item \textbf{Persona-fit}: Whether the user's behavior and language align well with their given profile.
\end{itemize}

Each dimension is rated on a 5-point Likert scale (1 = very poor, 5 = excellent). We additionally report an overall average score across all four dimensions (\textbf{All}), and compute inter-annotator agreement using Cohen’s Kappa to assess rating consistency.

\paragraph{Evaluation Results}

The results are shown in Table~\ref{tab:user_model_eval}. Among the evaluated models, \texttt{DeepSeek-V3} achieves the highest average score (4.23) across dimensions, along with the highest inter-annotator agreement ($\kappa=0.81$), indicating both high user simulation quality and rating reliability.

\begin{table*}[h]
\centering
\caption{Human evaluation results of user models (5-point scale).}
\label{tab:user_model_eval}
\begin{tabular}{lccccc|c}
\toprule
\textbf{Model} & \textbf{Coherence} & \textbf{Stability} & \textbf{Proactivity} & \textbf{Persona-fit} & \textbf{All (Avg.)} & \textbf{$\kappa$} \\
\midrule
GPT-4.1-mini    & 4.20 & 4.03 & \textbf{4.10} & 4.23 & 4.14 & 0.62 \\
GPT-4.1-nano    & 3.73 & 4.00 & 4.03 & 4.13 & 3.98 & 0.69 \\
DeepSeek-V3     & \textbf{4.37} & \textbf{4.27} & 4.07 & \textbf{4.33} & \textbf{4.26} & \textbf{0.71} \\
GPT-4o-mini     & 3.93 & 4.07 & 3.90 & 4.27 & 4.04 & 0.65 \\
\bottomrule
\end{tabular}
\end{table*}

\section{Reward Model Evaluation}
\label{app:reward_eval}

\subsection{Profile Reward Model}
\label{app:profile_reward_model}

\paragraph{Test Dataset Construction} 
To obtain the value of $|\hat{\mathcal{P}}_t \cap \mathcal{P}|$ in Equation~\ref{eq:profile_reward}, we employ a LLM to predict the number of semantically overlapping items between two profiles. To identify the most suitable prompt and LLM configuration for this task, we constructed a dedicated test dataset.

Given a reference profile, we create a rewritten version by modifying a subset of its items. Specifically, we randomly select $a$ items to be paraphrased—ensuring semantic equivalence while altering the surface form—and another disjoint set of $b$ items to be replaced with semantically different content. The final rewritten profile consists of the union of the $a$ paraphrased and $b$ altered items.

Each case includes the original and the rewritten profile, and the ground-truth overlap count $a$. We prompt LLM to predict the number of overlapping items, and compare output against the true value.

\paragraph{Human Evaluation of Dataset Quality} 
To assess the quality of the constructed dataset, we conducted a human evaluation on 300 randomly sampled profile item pairs, covering both \textit{Same-Meaning} and \textit{Different-Meaning} cases. Each pair consists of an original item and its rewritten counterpart, along with the system-assigned semantic label.

Three human annotators independently evaluate each pair and select one of the following judgments:
\begin{itemize}[leftmargin=*]
    \item \textbf{Same}: The items express the same meaning.
    \item \textbf{Different}: The items express different meanings.
    \item \textbf{Uncertain}: The semantic relationship is unclear or ambiguous.
\end{itemize}

A majority vote was used to determine the final label. If the system-assigned label matched the majority human judgment, the prediction was considered correct; otherwise—including \textit{Uncertain} cases—it was considered incorrect.

Table~\ref{tab:human_eval_results} reports the number of samples under each system label, the distribution of human judgments, and the corresponding accuracy.

\begin{table*}[h]
\centering
\caption{Human Evaluation Results on Profile Rewrite Dataset (300 samples)}
\label{tab:human_eval_results}
\begin{tabular}{lccccc}
\toprule
\textbf{System Label} & \textbf{Total} & \textbf{Same} & \textbf{Different} & \textbf{Uncertain} & \textbf{Accuracy (\%)} \\
\midrule
Same Meaning          & 150            & 138           & 6                  & 6                  & 92.0 \\
Different Meaning     & 150            & 9             & 134                & 7                  & 89.3 \\
\midrule
\textbf{Overall Accuracy} & 300       & -             & -                  & -                  & \textbf{90.7} \\
\bottomrule
\end{tabular}
\end{table*}

\paragraph{Evaluation of Profile Reward Model} 
We evaluate the LLM's ability to estimate profile overlap using the following metrics:
\begin{itemize}[leftmargin=*]
    \item \textbf{Exact Accuracy}: The proportion of predictions that exactly match the ground-truth overlap count $a$.
    \item \textbf{Fuzzy Accuracy}: The proportion of predictions with an absolute error $\leq 1$.
    \item \textbf{MSE and RMSE}: The mean squared error (MSE) and root mean squared error (RMSE) between predicted and ground-truth overlap counts, to quantify numerical deviation.
\end{itemize}

The prompt used for this task is shown in Figure \ref{fig:pr_prompt_zh}, with the English translation in Figure \ref{fig:pr_prompt_en}:

\begin{figure}[htbp]
    \centering
    \includegraphics[width=1\linewidth]{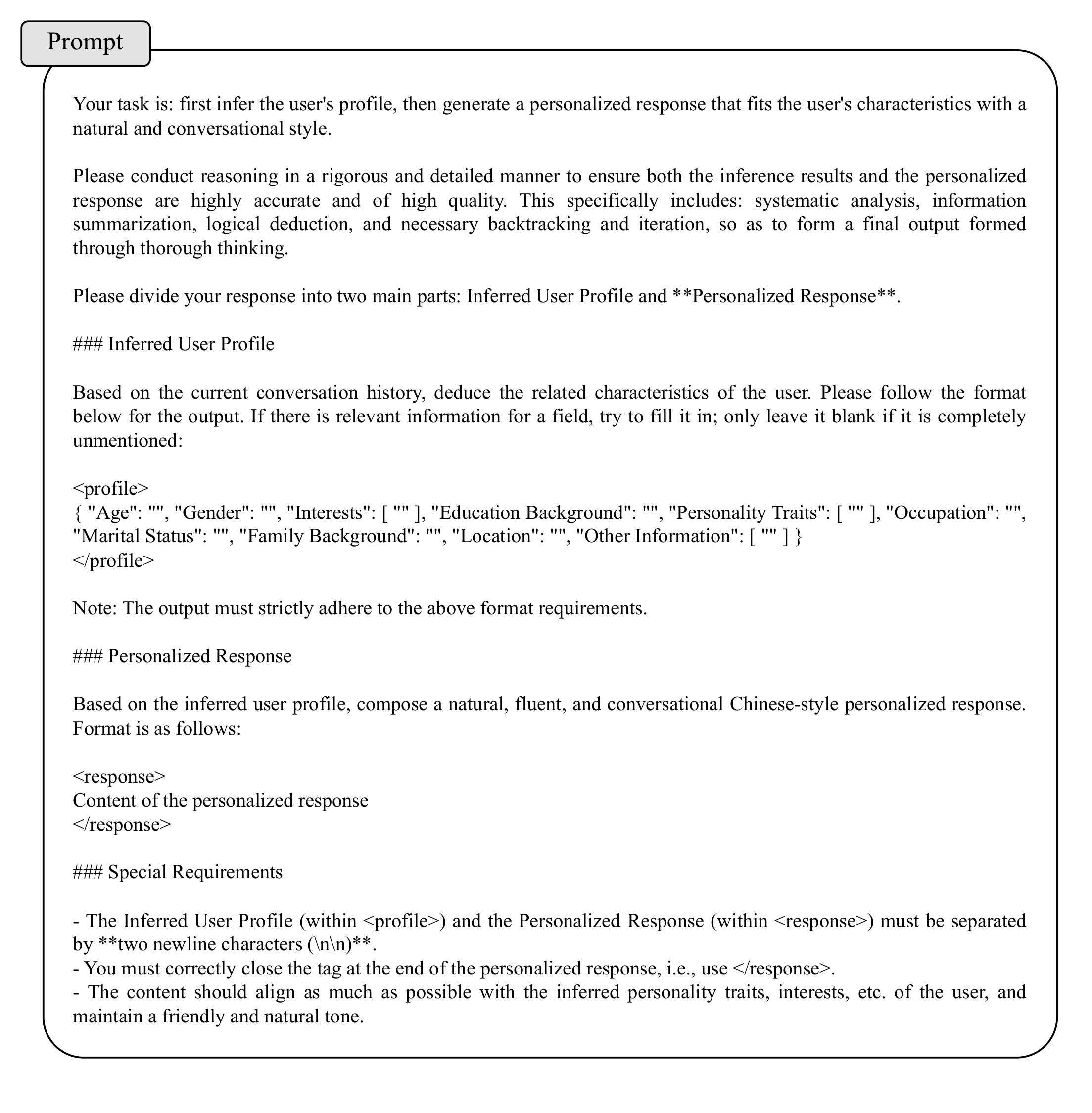}
    \caption{The prompt used for profile reward model (English Translation).}
    \label{fig:pr_prompt_en}
\end{figure}

\begin{figure}[htbp]
    \centering
    \includegraphics[width=1\linewidth]{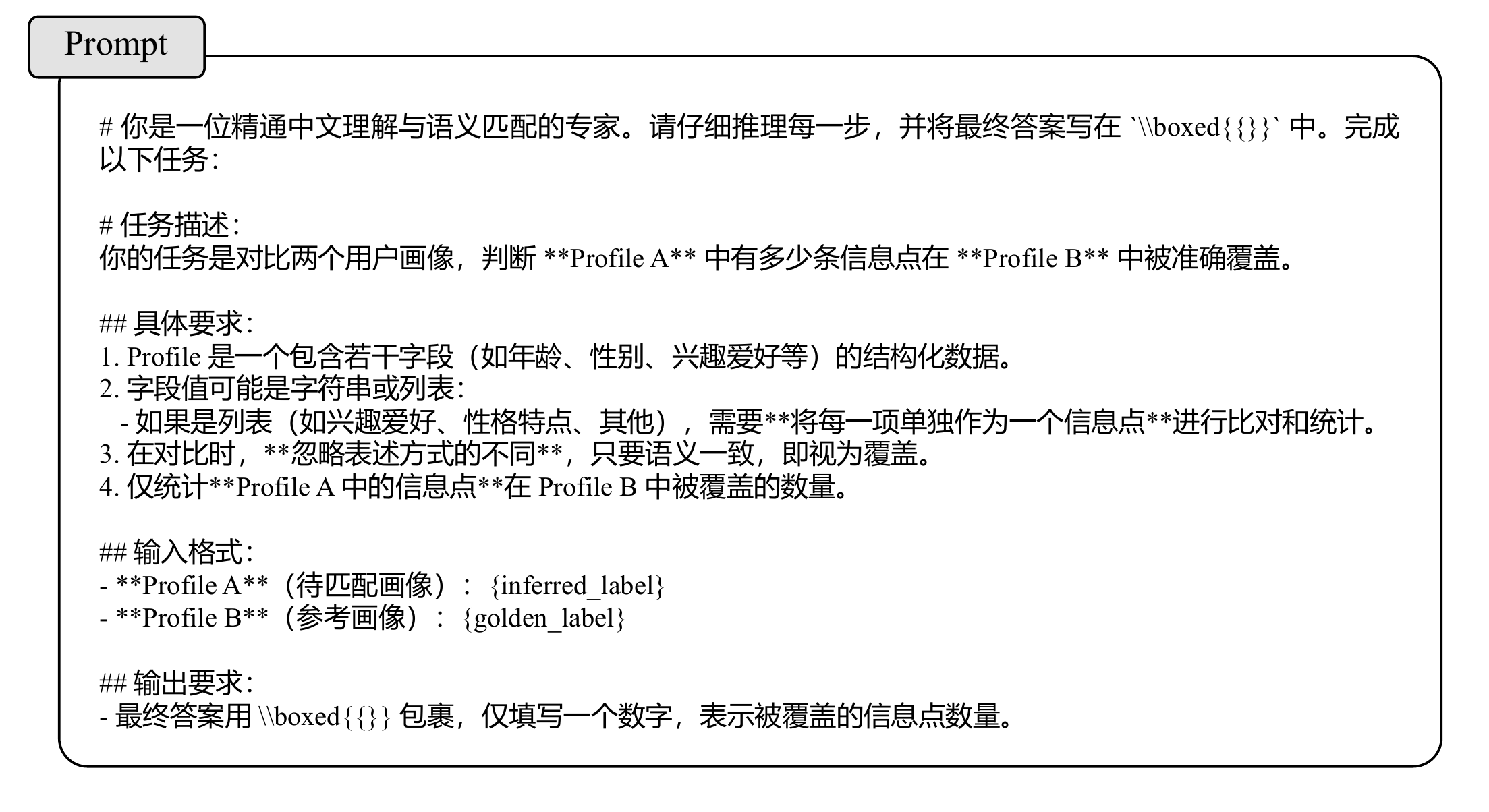}
    \caption{The prompt used for profile reward model.}
    \label{fig:pr_prompt_zh}
\end{figure}

Table~\ref{tab:profile_reward_eval} summarizes the results across different LLMs and prompt variants.

\begin{table*}[!ht]
\centering
\caption{Performance of different LLMs on the profile reward estimation task.}
\label{tab:profile_reward_eval}
\resizebox{0.6\linewidth}{!}{
    \begin{tabular}{lcccc}
    \toprule
    \textbf{Model} & \textbf{Exact Acc.} & \textbf{Fuzzy Acc.} & \textbf{MSE} & \textbf{RMSE} \\
    \midrule
    GPT-4.1-mini & 75.0 & 100.0 & 0.25 & 0.50 \\
    O3-mini      & 77.0 & 98.0  & 0.29 & 0.54 \\
    GPT-4.1-nano & 64.0 & 96.0  & 0.48 & 0.69 \\
    DeepSeek-V3  & 68.0 & 94.0  & 0.55 & 0.74 \\
    DeepSeek-R1  & 55.0 & 89.0  & 0.88 & 0.94 \\
    GPT-4o-mini  & 35.0 & 68.0  & 2.77 & 1.66 \\
    \bottomrule
    \end{tabular}
}
\end{table*}

\subsection{Response Reward Model}
\label{app:response_reward_model}

To evaluate the reliability of the response reward model during RL training, we performed a post-hoc human evaluation based on logged data. Specifically, we randomly sampled 300 data points from the training process, each consisting of an inferred profile $\hat{\mathcal{P}}_t$, a dialogue context, a generated response $r_t$, and the model's predicted scalar reward $R^{\text{response}}_t \in [0,1]$.

Each sampled instance was independently re-evaluated by human annotators using the same set of dimensions as the reward model—\textit{preference expression}, \textit{style consistency}, \textit{goal alignment}, \textit{persona coherence}, as well as the five binary quality criteria: \textit{naturalness}, \textit{relevance}, \textit{logical consistency}, \textit{engagement}, and \textit{informativeness}. The final human reward label was computed as:
\[
R^{\text{human}} = N \cdot R \cdot L \cdot G \cdot F \in \{0, 1\}
\]

We then construct a confusion matrix between the human labels and model predictions, and calculate classification metrics including accuracy and F1 score. The results are shown in Table~\ref{tab:response_confusion}.

We computed the confusion matrix between the model predictions and human annotations, as well as standard evaluation metrics, including Cohen’s Kappa for inter-rater agreement. Results are shown in Table~\ref{tab:response_confusion} and Table~\ref{tab:response_metrics}.

\begin{table*}[h]
\centering
\caption{Confusion matrix of response reward model (300 samples)}
\label{tab:response_confusion}
\begin{tabular}{lcc}
\toprule
\textbf{} & \textbf{Human: 1} & \textbf{Human: 0} \\
\midrule
\textbf{Model: 1} & 124 & 21 \\
\textbf{Model: 0} & 18 & 137 \\
\bottomrule
\end{tabular}
\end{table*}

\vspace{0.5em}

\begin{table*}[h]
\centering
\caption{Evaluation metrics of response reward model}
\label{tab:response_metrics}
\begin{tabular}{lcccccc}
\toprule
\textbf{Accuracy} & \textbf{Precision} & \textbf{Recall} & \textbf{F1 Score} & \textbf{Specificity} & \textbf{Cohen’s Kappa} \\
\midrule
0.87 & 0.855 & 0.873 & 0.864 & 0.867 & 0.740 \\
\bottomrule
\end{tabular}
\end{table*}

The model demonstrates strong agreement with human judgment across all dimensions, achieving an accuracy of 87\% and a Cohen’s Kappa of 0.74, indicating substantial consistency. These results validate the reward model’s reliability as a reinforcement signal during training.

\begin{figure}
    \centering
    \includegraphics[width=1\linewidth]{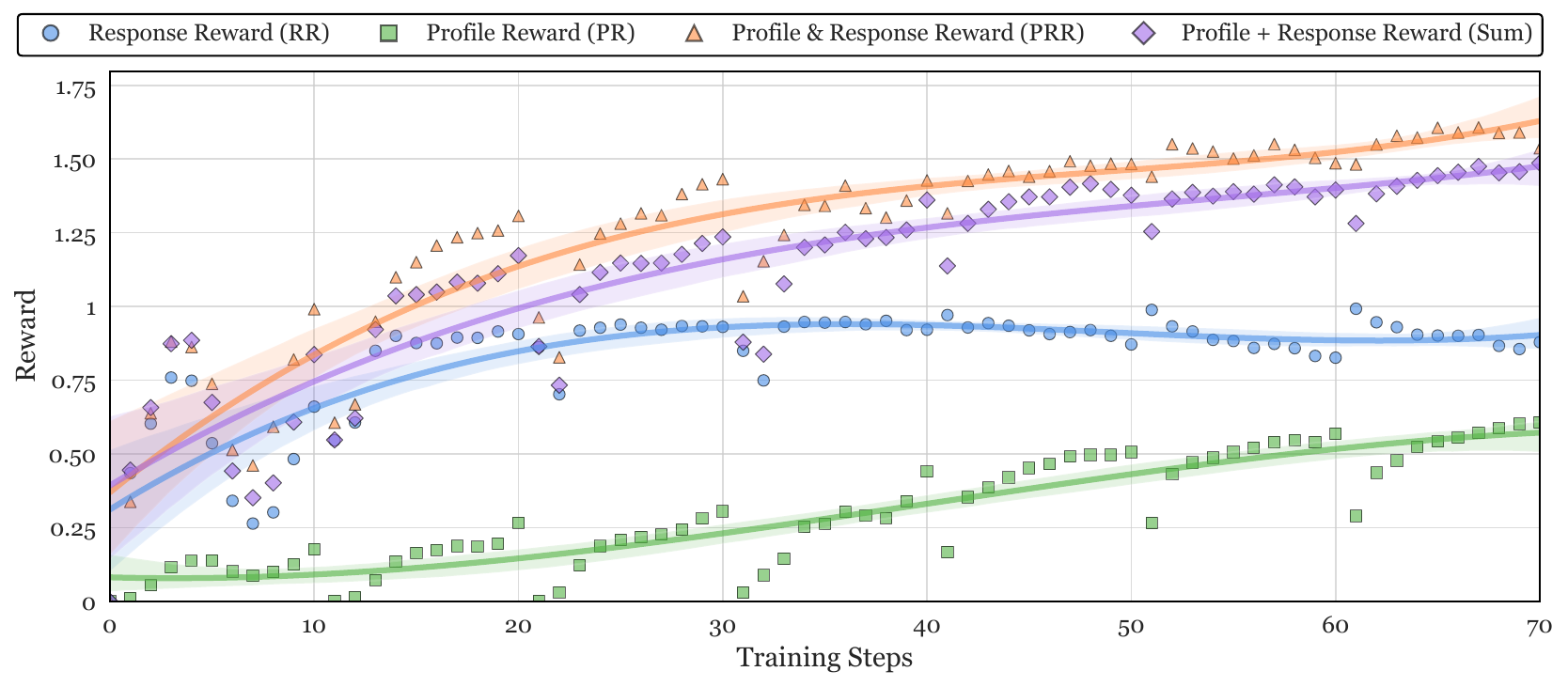}
    \caption{Reward progression across training steps for different reward configurations.}
    \label{fig:reward_curve}
\end{figure}

\section{Reward Curve}
\label{app:reward_curve}

In the section, we present the reward curves depicted in Figure \ref{fig:reward_curve}, which illustrate the performance of our proposed Profile \& Response Reward (PRR) method compared to the traditional Response Reward (RR) and Profile Reward (PR) methods. The results unequivocally demonstrate that the PRR approach outperforms both RR and PR across various training steps.
A closer examination of the PRR curve reveals a consistent upward trend, indicating superior reward accumulation over time. Notably, when comparing PRR with the sum of individual rewards from Profile + Response (denoted as "Sum"), it becomes evident that integrating rewards from both profile and response mechanisms fosters a synergistic effect. This synergy enables mutual enhancement between profile and response components, leading to more effective and efficient learning outcomes.
The observed trends underscore the importance of considering both profile and response dimensions simultaneously in the reward structure. By doing so, the PRR method not only achieves higher overall rewards but also promotes a balanced and comprehensive optimization process. These findings highlight the potential of the RLPA framework in enhancing the performance of reinforcement learning models, particularly in scenarios where profile and response interactions play a crucial role.


\end{document}